\documentclass{article}

\usepackage{CustomCommands}
\begin{document}
\title{Demonstration of effective UCB-based routing in skill-based queues on real-world data}
\author{Sanne van Kempen, Jaron Sanders, Fiona Sloothaak, Maarten G. Wolf}
\date{}
\maketitle

\begin{abstract}
This paper is about optimally controlling skill-based queueing systems such as data centers, cloud computing networks, and service systems.
By means of a case study using a real-world data set, we investigate the practical implementation of a recently developed reinforcement learning algorithm for optimal customer routing. 
Our experiments show that the algorithm efficiently learns and adapts to changing environments and outperforms static benchmark policies, indicating its potential for live implementation. 
We also augment the real-world applicability of this algorithm by introducing a new heuristic routing rule to reduce delays. 
Moreover, we show that the algorithm can optimize for multiple objectives: next to payoff maximization, secondary objectives such as server load fairness and customer waiting time reduction can be incorporated. 
Tuning parameters are used for balancing inherent performance trade--offs. 
Lastly,  we investigate the sensitivity to estimation errors and parameter tuning, providing valuable insights for implementing adaptive routing algorithms in complex real-world queueing systems.

\end{abstract}

\section{Introduction}\label{sec:introduction}
In skill-based queues, different customer types request service from a set of heterogeneous servers, and compatibility relations determine which customer--server matches are allowed. 
Since such systems are widely used in data centers, cloud computing networks, healthcare applications, and service systems, efficient resource allocation is highly valuable~\cite{Chen2020}. 
Full theoretical understanding of such systems, however, is challenging due to the complex queueing dynamics and compatibility relations \cite{Adan2012,Adan2014,Weiss2021}. \\

Canonical routing policies such as (i) \gls{FCFS}---\gls{ALIS}, (ii) the $c\mu$ rule, (iii) threshold-based policies, or (iv) Serve--the--Longest--Queue policy are typically static by design \cite{Adan2014, Smith1956}.
Although such policies are easy to implement, they ignore long-term effects and/or stability constraints.
Moreover, they offer limited robustness and adaptability to optimize for various objectives such as maximizing the quality of customer--server matches, waiting time, fairness, and server loads. \\

To address these challenges, there has been a growing interest in adaptive control policies that make data-driven decisions based on real-time feedback~\cite{Zhong2022,Krishnasamy2018b,Shah2020}. 
In particular, recent works underline the potential of machine learning algorithms for practical implementation in queueing systems.
For example, 
\cite{Dai2022} applies a proximal policy optimization for routing in a multiclass queueing networks, 
\cite{Tessler2022} optimizes congestion control in a data center using a reinforcement learning algorithm, 
\cite{Liang2020} introduces an actor--critic method for job scheduling in a multi-resource job scheduling, 
and \cite{Comte2023} introduces new score-aware gradient estimators which are applied in policy-gradient methods for stochastic optimal control problems.\\

Motivated by this, we focus on a recently developed utility-maximizing reinforcement learning algorithm, called ``\gls{UCBQR}'', for routing in skill-based queues~\cite{Kempen2024}. 
This algorithm integrates multi-armed bandit techniques and \gls{UCB} estimators in skill-based queueing systems~\mbox{\cite{Auer2002a,Lattimore2020}} to make routing decisions under uncertainty. 
It is shown that the algorithm is nearly optimal in the sense that its regret accumulation matches the asymptotic lower bound up to logarithmic terms.
While the theoretical performance of \gls{UCBQR} has been  analyzed in detail, its practical value in real-world queueing systems was not yet demonstrated. 
For example, queue lengths, delays, and fairness aspects of the algorithm are not considered in~\cite{Kempen2024}.
To this end, we perform an extensive numerical study of the potential applications and extensions for practical use of the algorithm in a realistic setting.
We adapt the algorithm for implementation in a practical setting: 
this includes action space reduction and dealing with temporary system instability and time-varying arrival rates and service capacity. 
While these phenomena are often neglected in synthetic experiments, they are critical for realistic applications. \\

To prepare for potential implementation of the algorithm in a real-life setting, we provide with this paper a simulation framework designed to accommodate for generic queueing system configurations. 
This includes the feature to set arbitrary compatibility relations and general arrival and service processes, making it a valuable tool for testing performance and robustness of learning-based routing policies in complex realistic environments.
The simulation is made publicly available\footnote{See \href{https://github.com/SannevanKempen/UCBQR}{https://github.com/SannevanKempen/UCBQR}.}. 
To demonstrate the practical use of the \gls{UCBQR} routing algorithm, we draw on data from a large-scale call center of a U.S.\ bank, made publicly available by the \gls{SEELAB}\footnote{See \href{https://seelab.net.technion.ac.il/data/}{https://seelab.net.technion.ac.il/data/}.}. \\

We demonstrate that the modified \gls{UCBQR} routing algorithm (\Cref{alg:learning_alg}) is well-suited for implementation in a large, realistic queueing system, even when the environment is subject to change. 
We focus on payoff maximization, where payoff models the quality of customer--server routings and is defined as the empirical ratio of successful service completions. 
We show that the \gls{UCBQR} algorithm successfully accumulates high payoff while learning the payoff parameters and estimating the (time-varying) arrival and service rates.
In one representative scenario, our algorithm closes the optimality gap by over 70\% when compared to naive random routing (see \Cref{fig:reward_June2001}). 
By also developing a new heuristic routing rule as a new subroutine for \gls{UCBQR}, customer waiting times can be reduced significantly. 
Moreover, we show that multiple objectives can be optimized simultaneously, with tuning parameters balancing the trade--offs. 
Lastly, the algorithm is not black-box nor computationally heavy: the algorithm's runtime is comparable with those of other routing policies like \gls{FCFS}---\gls{ALIS} or random routing.

\subsection{Experiment overview}
To showcase the potential of the algorithm in real-world settings, we conduct a series of experiments, which we now briefly discuss.
With this case study we achieve three main objectives, namely 
(i) implementation of a state--of--the--art reinforcement learning  algorithm in a realistic large-scale service system (Experiment 1), 
(ii) extension and improvement of the algorithm to address practical objectives (Experiments 2--4), 
and (iii) evaluation of the adaptability and robustness of the algorithm (Experiments 5, 6). 
Along the way, we discuss insightful numerical results that can be of interest for future studies. 
\begin{enumerate}
    \item {\bf Proof of concept.} 
    We find that the \gls{UCBQR} algorithm consistently outperforms the benchmark policies in terms of payoff accumulation, including a ``greedy'' policy and a variant of the $c\mu$ rule that is modified for payoff maximization~\cite{Smith1956}.
    The loss in payoff due to learning the payoff parameters and estimating the (time-varying) arrival and service rates is at most 2\% when compared to the all-knowing payoff maximizing ``oracle'' policy.  
    \item {\bf Alternative routing rule to decrease waiting times.} 
    As our proposed \gls{UCBQR} routing algorithm primarily aims for payoff maximization, it typically leads to higher customer delays than benchmark non-adaptive routing policies. 
    These waiting times can be improved by noting that the routing rule used in \gls{UCBQR} dispatches customers to servers according to prescribed routing rates (\Cref{alg:routing}). By replacing this subroutine by a new heuristic routing rule, we show that the waiting times can be reduced to the levels of more delay-friendly policies (e.g.\ \gls{FCFS}---\gls{ALIS}), while still maintaining high payoff accumulation. 
    This heuristic routing rule (\Cref{alg:tree}) is inspired by the \gls{JSQK} customer dispatching policy presented in~\cite{Fu2022}.
    \item {\bf Combining different performance objectives.}
    When managing queues in practice, payoff maximization is rarely the only performance measure. 
    Instead, other measures like customer waiting time, fairness, and server load can be of interest. 
    We show how \gls{UCBQR} can be adapted to accommodate for multiple, possibly competing objective functions. 
    A balance in the trade--off between the different objectives can be made by tuning a set of system parameters. 
    \item {\bf Resilience in bursty scenarios.}  
    In real-life scenarios, the customer arrival process can be affected by incidents, outages, or special events that lead to customer surges or bursty arrival processes. 
    To demonstrate the robustness of our algorithm under such events, we simulate an incident scenario by adding artificial customer bursts on top of the regular arrival process from the data set. 
    We observe that our learning-based algorithm achieves overall the highest cumulative payoff compared to the benchmark policies. 
    Moreover, it recovers from the arrival surges in terms of waiting times, although it takes longer than the benchmark policies. 
    This showcases the spirit of our approach: despite short-term surges, our algorithm focuses on the long-term optimization of customer---server matching quality.
    \item {\bf Robustness against estimation errors.}
    Instead of assuming fixed and known arrival and service rates, the \gls{UCBQR} algorithm uses online estimators.  
    We investigate in what way the accuracy of the estimators influences the algorithm's performance. 
    Our results indicate that empirical estimators work sufficiently well, and relatively little performance benefit can be gained by using the true parameter values for the arrival and service rate. 
    \item {\bf Balancing reactivity and complexity.}
    \gls{UCBQR} learns the payoff parameters in an episodic structure: 
    the payoff, arrival rate, and service rate estimators are updated at the start of each episode using observed samples from previous episodes. 
    The episode length therefore controls the learning rate: the shorter the episodes, the faster the payoff parameters are learned. 
    However, there exists a trade--off with computational complexity, since computations need to be performed at each episode. 
    The episode length should therefore be small enough to capture the changes in the time-varying customer arrival rate, since otherwise customer waiting times become large.
    In our experimental setting, we find that the effect of the episode length on payoff accumulation is small. 
\end{enumerate}

\subsection{Paper outline}
The modified \gls{UCBQR} algorithm is presented in \Cref{sec:algorithm}. 
Implementation challenges are discussed in \Cref{sec:sim_framework}, and experimental results are discussed in \Cref{sec:experiments}.
Lastly, we conclude briefly in \Cref{sec:conclusion}.

\section{Algorithm description}\label{sec:algorithm}
We consider a skill-based queueing system with customer types $\calI = \{1,\dots,I\}$, servers $\calJ = \{1,\dots,J\}$, and a set of compatibility lines connecting customer types and servers $\calL = \{(ij): i\in\calI, j\in\calJ\}$. 
Customers wait in the queue of their type until they start service at one of the compatible servers. 
Each service completion of a type-$i$ customer at server $j$ generates a random Bernoulli payoff with mean $\theta_{ij}\in[0,1]$. \\

We present a modified version of the routing algorithm developed by~\cite{Kempen2024}, called \gls{UCBQR}, in \Cref{alg:learning_alg}.
The goal of \gls{UCBQR} is to maximize the total average payoff $\sum_{(ij)\in\calL} \theta_{ij}\E[D_{ij}(t)]$, where $\E[D_{ij}(t)]$ is the number of type-$(ij)$ departures,  while learning the payoff parameter $\theta_{ij}$ and maintaining queue stability. 
\Cref{alg:learning_alg} considers the following optimal transport \gls{LP} that optimizes the routing rates $x_{ij}$ of type-$i$ customers to server $j$ to maximize the total average payoff~\eqref{eq:LP_eps_obj}.
Constraint~\eqref{eq:LP_eps_constr_x} guarantees that all customers are routed to some server, and constraints~\eqref{eq:LP_eps_constr_mu} restricts that the total server load $\sum_{ij} x_{ij}/\mu_{ij}$ does not exceed $1-\eps$ for some $\eps\in(0,1)$, ensuring stability of the queueing system. 
\begin{subequations}
    \begin{align}
        \LP{\lambda, \mu, \theta,\eps}: \ \ \max_{x\geq 0} \ \ &\sum_{(ij)\in\calL} \theta_{ij} x_{ij}, \label{eq:LP_eps_obj} \\
        \textrm{s.t.} \ \ &\sum_{j\in\calS_i} x_{ij} = \lambda_i, \ \ \ \forall i\in\calI, \label{eq:LP_eps_constr_x}\\
        &\sum_{i\in\calC_j} \frac{x_{ij}}{\mu_{ij}} \leq 1 - \eps, \ \ \ \forall j\in\calJ. \label{eq:LP_eps_constr_mu} 
    \end{align}
    \label{eq:LP_eps}
\end{subequations}

\Cref{alg:learning_alg}  divides the time horizon into episodes. 
At the start of episode $k$, the estimators for (i) the average payoffs $\hat\theta_{ij}^k$, (ii) the arrival rates $\hat\lambda_i^k$, and (iii) the service rates $\hat\mu_{ij}^k$ are updated using collected samples. 
For the payoff parameters, \gls{UCB} estimators~\mbox{\cite{Auer2002a,Lattimore2020}} are used which are updated in~\eqref{eq:UCB_update_algo_U_per_line}. 
Consequently, the set of routing rates $x_{ij}$, $(ij)\in\calL$ are chosen such that the value of $\LP{\hat\lambda^k,\hat\mu^k,\hat\theta^k,\eps}$ is maximized. 
During the episode, customers are then routed to servers according to the rates $x$. 
To implement this routing rule, virtual queues are used (\Cref{alg:routing}).\\

\Cref{alg:learning_alg} is an adaptation of the routing algorithm introduced in~\cite{Kempen2024}. 
Inherent to real-life situations, not all assumptions made in \cite{Kempen2024} are satisfied in practical systems. 
We briefly discuss the most significant changes to make the algorithm suitable for practical implementation. 

\begin{itemize}
    \item {\bf Action space reduction.} 
    The algorithm described in~\cite{Kempen2024} considers the set of all \glspl{BFS} of the \gls{LP} as its action space, i.e., the set of candidate solutions. 
    Each episode, the algorithm chooses the solution (action) with the highest estimated payoff based on the \gls{UCB} estimators $\hat\theta$.
    Since the number of \glspl{BFS} scales exponentially with the number of customer types $I$, servers $J$, and lines $|\calL|$, this approach can be computationally prohibitive for large queueing systems. 
    To reduce the complexity of the algorithm, we instead solve~\eqref{eq:LP_eps} directly at the start of each episode using current payoff estimates, and use its maximizer as the desired routing rates (\Cref{line:solve_LP_eps}).
    This approach avoids the need to enumerate or store the full set of \glspl{BFS}. 
    For fixed values $\lambda, \mu, \theta,\eps$, \eqref{eq:LP_eps} can be solved efficiently by modern solvers that rely on, e.g., the simplex method or interior-point methods~\cite{Bertsimas1997}, making this a scalable alternative.
    \item {\bf Stability.}
    In~\cite{Kempen2024}, a notion of stability is assumed, which assures that~\eqref{eq:LP_eps} is feasible. 
    However,~\eqref{eq:LP_eps} can be infeasible if the arrival volume of certain customer types exceeds the service capacity of its compatible servers. 
    In practice, such situations are to be expected in real-life queueing systems, where temporary overloads can occur. 
    In addition, estimation errors in $\hat{\lambda}$ and $\hat{\mu}$ can also lead to infeasibility of \eqref{eq:LP_eps}. 
    In case \eqref{eq:LP_eps} is infeasible under the estimated $\hat{\lambda}$ and $\hat{\mu}$, we consider the following alternative \gls{LP}:
    \begin{subequations}
        \begin{align}
            \LPa{\lambda, \mu, \theta,\eps}: \ \ \max_{x\geq 0} \ \ &\sum_{(ij)\in\calL} \theta_{ij} x_{ij} - \sum_{i\in\calI} p x_{iz}, \\
            \textrm{s.t.} \ \ &\sum_{j\in\calS_i} x_{ij} + x_{iz} = \lambda_i, \ \ \ \forall i\in\calI, \\
            &\sum_{i\in\calC_j} \frac{x_{ij}}{\mu_{ij}} \leq 1 - \eps, \ \ \ \forall j\in\calJ.  
        \end{align}
        \label{eq:LP_adapted}
    \end{subequations}
    $\textrm{LP}^{\textrm{a}}$  introduces a new fictitious ``rejection'' server $z$ with infinite capacity and connections to all customer types. 
    Part of the arrival rate $\lambda_i$ of customer $i$ can be routed to server $z$, which is managed by the decision variables $x_{iz}$, but this is penalized in the objective by a negative payoff $p$. 
    If the penalty $p$ is large, then weight is given to $x_{iz}$ only when there is no other feasible solution. 
    This modified \gls{LP} is always feasible and provides a robust alternative when \eqref{eq:LP_eps} is infeasible. 
    We note that the fictitious rejection server is not actually used when making routing decisions: after we obtain the optimal solution $x$ from~\eqref{eq:LP_adapted}, we compute routing probabilities $p_{ij}$ by normalizing the rates according to $p_{ij} = x_{ij}/\sum_{k\in\calI} x_{ik}$, where server $z$ is excluded (see~\Cref{line:routing_prob} in \Cref{alg:learning_alg}). 
    \item {\bf Arrival and service processes.} 
    In~\cite{Kempen2024}, type-$i$ customers are assumed to arrive according to independent Poisson processes with known rates $\lambda_i >0$, and service times are assumed to be independent and exponentially distributed with known rates $\mu_j>0$. 
    Since there is uncertainty and time-varying behavior in arrival rates and service capacity in real-life queueing systems, we use estimators (\Cref{line:lam_mu_est}). 
    This introduces more noise in the estimation errors that needs to be dealt with efficiently. 
    The estimators $\hat\lambda$ and $\hat\mu$ are discussed in more detail in \Cref{sec:arrival_estimator} and \Cref{sec:service_estimator}, respectively. 
\end{itemize}

The pseudocode for our modified \gls{UCBQR} algorithm is shown in \Cref{alg:learning_alg} below.

\begin{algorithm}[H]
    \caption{Modified UCB--Queue Routing (\gls{UCBQR}).}
    \label{alg:learning_alg}
    \begin{algorithmic}[1]
        \State Initialize $k=1$. For all $(ij)\in\calL$, initialize $T_{ij}(0) = \bar{\theta}_{ij}(0) = 0$ and $\hat{\theta}_{ij}(0) = \infty$.
        \For{each episode $k=1,2,\dots$ of length $h$} \label{line:episode_length}
            \State Set $\hat\lambda^k$ according to~\eqref{eq:arr_est} and $\hat\mu^k$ according to~\eqref{eq:serv_est}. \label{line:lam_mu_est} 
            \If{\LP{\hat{\lambda}^k,\hat{\mu}^k,\hat{\theta}^k,\eps}~\eqref{eq:LP_eps} if feasible}  \Comment{Determine routing rates.}
                \State  Solve \LP{\hat{\lambda}^k,\hat{\mu}^k,\hat{\theta}^k,\eps} in~\eqref{eq:LP_eps}; denote its maximizer by $\hat{x}^k$. \label{line:solve_LP_eps}
            \Else
                \State Solve \LPa{\hat{\lambda}^k,\hat{\mu}^k,\hat{\theta}^k,\eps} in~\eqref{eq:LP_adapted}; denote its maximizer by $\hat{x}^k$.  \label{line:solve_LP_adapted}
            \EndIf
            \State Serve customers according to $\hat{x}^k$ by applying Algorithm~\ref{alg:routing}: FCFS--RR$(\hat{x}^k, h)$.  \Comment{Route customers.}
            \State For each line $(ij)\in\calL$, observe $N_{ij}^k\in\N_{\geq 0}$ payoff samples $(Y_{ij}^\ell)_{\ell=1,\dots,N_{ij}^k}$. \label{line:observe_payoffs}
            \State For each line $(ij)\in\calL$, update \label{line:update_ucb} \Comment{Update estimators based on samples.} 
            \begin{align}
                T_{ij}(k)
                &= T_{ij}(k-1) + N_{ij}^k, 
                \label{eq:UCB_update_algo_T}
                \\
                \bar{\theta}_{ij}(k)
                &= \frac{\bar{\theta}_{ij}(k-1) T_{ij}(k-1) + \sum_{\ell=1}^{N_{ij}^k} Y_{ij}^\ell}{T_{ij}(k)}, 
                \label{eq:UCB_update_algo_theta_bar}
                \\
                \hat{\theta}_{ij}^k &=
                    \begin{cases}
                        \bar{\theta}_{ij}(k) + \sqrt{\frac{ \ln(k)}{T_{ij} (k)}} &\textrm{if} \  T_{ij}(k) > 0, \\
                        \infty &\textrm{if} \ T_{ij}(k) = 0.
                    \end{cases}
                \label{eq:UCB_update_algo_U_per_line}
            \end{align}
        \EndFor
    \end{algorithmic}
\end{algorithm}

\begin{algorithm}[H]
    \caption{FCFS--RR$(x, h)$.}
    \label{alg:routing}
    \begin{algorithmic}[1]
        \State Set $p_{ij} = x_{ij} / \sum_{k\in\calJ} x_{ik}$ for all $(ij)\in\calL$.  \label{line:routing_prob} \Comment{Compute routing probabilities.}
        \State Empty all virtual queues. \label{line:reallocation0} \Comment{Shuffle virtual queues.} 
            \State For each waiting type-$i$ customer, assign it to the virtual queue of server $j$ with probability $p_{ij}$.  \label{line:reallocation1} 
            \State For each $j\in\calJ$, order the customers in virtual queue $j$ by their arrival time.
            \label{line:reallocation2}
        \For{time $t = 0,\dots, h$} \Comment{Route customers.} 
        \For{each arrival of a type-$i$ customer at time $t$}
            \State Assign customer to the virtual queue of server $j$ with probability $p_{ij}$.
            \State If the server is idle, start service. 
        \EndFor
        \For{each service completion at server $j$ at time $t$}
            \State Obtain a payoff $Y_{ij}\sim\bern{\theta_{ij}}$, where $i$ is the departing customer's type.
            \State If the virtual queue of server $j$ is non-empty, start service of the customer at the head of the queue.
        \EndFor
        \EndFor
    \end{algorithmic}
\end{algorithm}

\section{Algorithm implementation}\label{sec:sim_framework}
In this section we discuss the challenges that we encountered while implementing \Cref{alg:learning_alg}. 
Although these preparatory steps are required for implementation of the algorithm, 
it is not strictly necessary to gain full understanding before reading the main results in \cref{sec:experiments}.
The data set, necessary data preparation and a brief overview of the simulation setup are discussed in \ref{app:implementation}.

\subsection{Implementation challenges}
We split the data into training and testing data. 
The first two months of the data set, April and May 2001, will be used as training set to tune model parameters. 
All other months are used as testing data for the experiments in \Cref{sec:experiments}.

\subsubsection{Defining payoff parameters}
To apply \Cref{alg:learning_alg}, we need to define payoff parameters $\theta_{ij}$ for the compatible lines $(ij)$.
Since there are no measures of costs or customer satisfaction scores associated with this data set, we instead define a payoff based on service outcomes. 
In the data set, each call has an outcome label, including ``transfer'', where a customer is transferred to an agent of another agent group, and ``conference'', where the agent consults a second agent or supervisor.
In both cases, the outcome indicates that additional resources were necessary to handle the call. 
We therefore define $\theta_{ij}$ as the ratio of successful type-$(ij)$ services, i.e., services that do not end in transfer or a conference. 
Concretely, for a particular day $d$ we let $\theta_{ij}^d := s_{ij}^d / n_{ij}^d$ where $s_{ij}^d$ is the number of type-$i$ calls that are routed to agent $j$ on day $d$ in the data set and do not end in transfer nor a conference, and $n_{ij}^d$ is the total number of $(ij)$ calls on day $d$. 
Note that the value of $\theta$ thus depends on the day. 

\subsubsection{Time-varying agent populations}
In the data set, each agent group consists of several agents (see \Cref{tab:agent_count} for an example).
For simplicity, we use one server for each agent group in the simulation. 
The service speed of each server is then scaled proportionally to the number of working agents in the data. 
This approximation can be considered similar to approximating an $M(\lambda)/M(\mu)/n$ queueing system by an $M(\lambda)/M(n\mu)/1$ system. 
Between these systems, the stationary probabilities are very close for states with $n$ or more customers~\cite{Cohen1981}. 
The data set contains the work hours of individual agents, including the duration spent answering calls or being available. 
For simplicity, we choose to update the number of available agents in the simulation every hour.
For each hour, we obtain from the data the total number of agents per group that worked anytime during that hour. 

\begin{minipage}{.15\textwidth}
    \centering
    \begin{table}[H]
        \centering
        \adjustbox{width=1\linewidth}{
        \begin{tabular}{cc}
            \toprule 
            Agent group & Agents \\ 
            \midrule
            1 & 199 \\
            2 & 192 \\
            6 & 33 \\
            8 & 35 \\
            9 & 43 \\
            11 & 42 \\
            14 & 21 \\
            18 & 67 \\
            $\vdots$ &$\vdots$ \\
            \bottomrule
        \end{tabular}
        }
        \caption{Number of working agents per agent group on Tuesday May 1, 2001.}
        \label{tab:agent_count}
    \end{table}
\end{minipage}
\hspace{.5em}
\begin{minipage}{.32\textwidth}
        \begin{figure}[H]
            \centering
            \includegraphics[width=\linewidth]{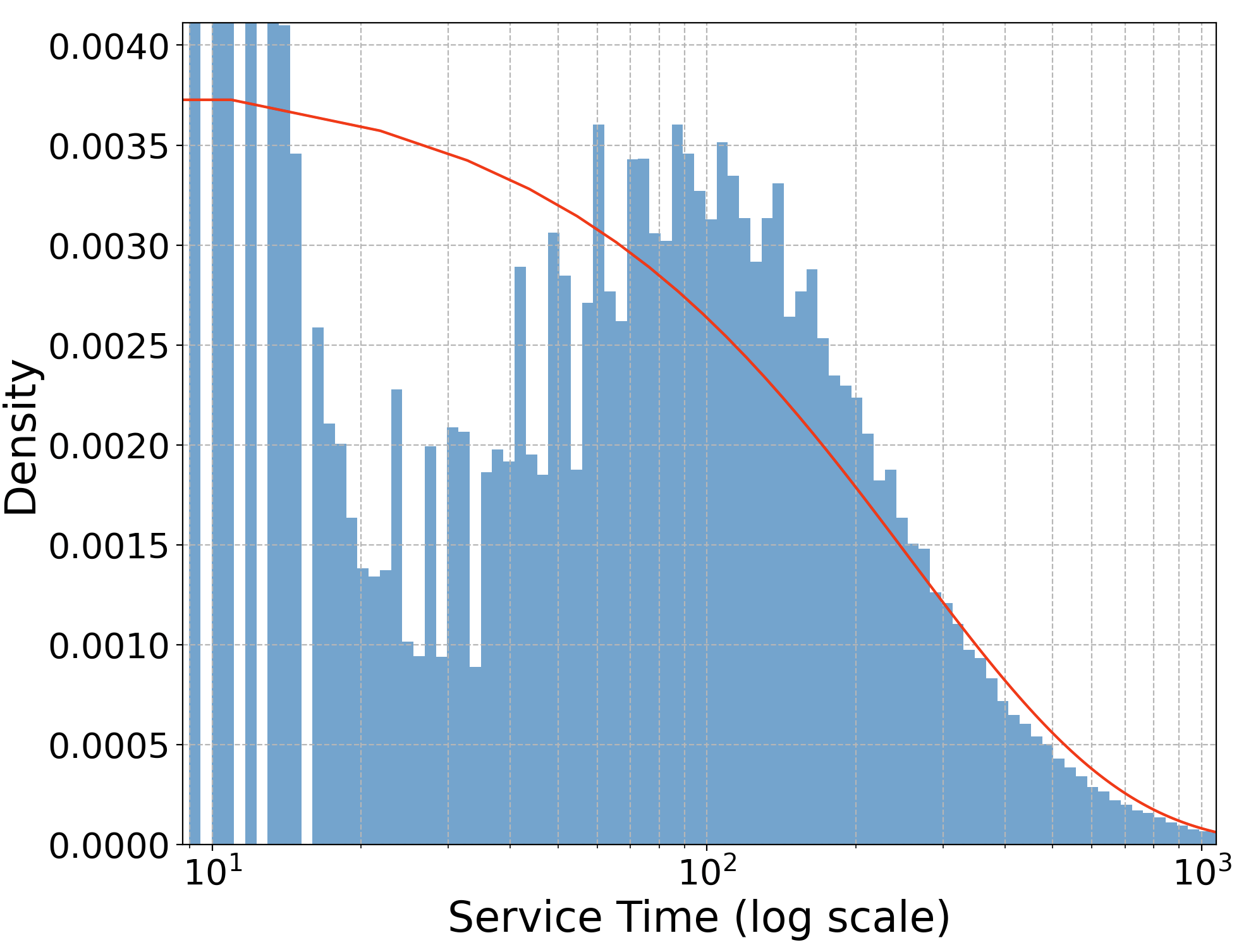}
            \caption{Histogram of type-$(11)$ service times in May 2001 based on 252,103 samples, along with the density of an exponential distribution with parameter 257.}
            \label{fig:servtimes_data_May2001}
            \end{figure}
\end{minipage}
\hspace{.5em}
\begin{minipage}{.46\textwidth}
        \begin{figure}[H]
            \centering
            \includegraphics[width=\linewidth]{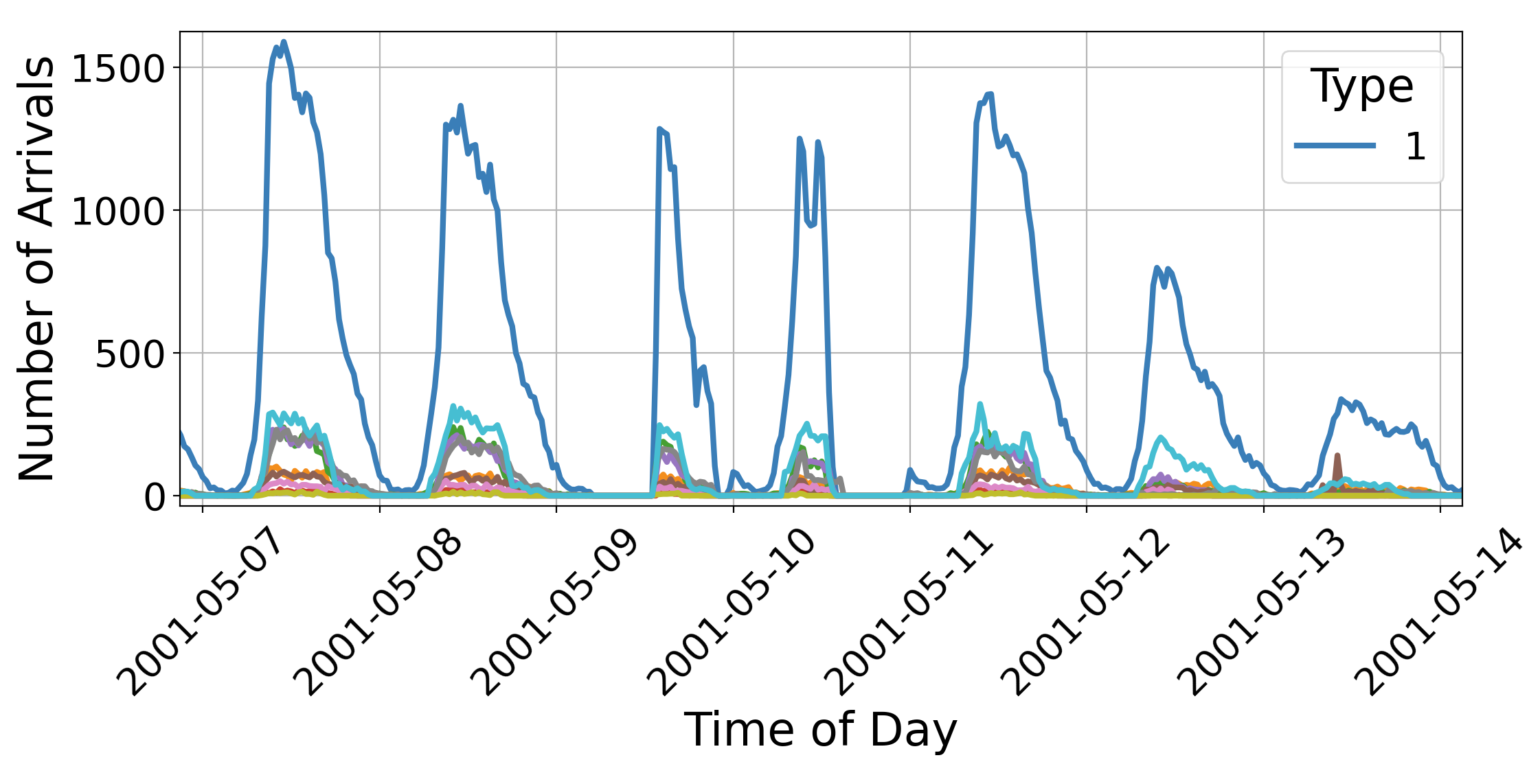}
            \caption{Number of customer arrivals accumulated per 30-minute intervals for the first week of May 2001. 
            The call center is closed on the morning of May 9 and the afternoon of May 10.}
            \label{fig:arrivals_May2001}
        \end{figure}
\end{minipage}

\subsubsection{Service time distribution}\label{sec:service_estimator}
In \Cref{alg:learning_alg} we have to specify the service rate $\hat\mu_{ij}^k$ for all lines $(ij)\in\calL$. 
\cite{Kempen2024} assumes independent and exponentially distributed service times, but the service times in the data are unlikely to be actually distributed as such. 
For example, \Cref{fig:servtimes_data_May2001} shows the empirical distribution of service times of type-1 customers at agent group 1 in May 2001, along with the best fitting exponential distribution. 
Here, ``best'' is measured in terms of the \gls{AIC}. \par \vspace{1em}
In order to make the simulation as close to the reality as possible, the service times of \mbox{type-$i$} customers at server $j$ in the simulation are sampled from the empirical distribution of the data set from that month. 
We define the service rate estimator $\hat\mu_{ij}^k$ in \Cref{line:lam_mu_est} as the inverse of the empirical mean durations of type-$(ij)$ services that have been completed before the start of episode $k$, i.e., 
\begin{align}\label{eq:serv_est}
    \hat{\mu}_{ij}^k :=\frac{M_k}{\sum_{\ell=1}^{M_k} m_{ij}^\ell},    
\end{align}
where $M_k$ is the number of type-$(ij)$ service completions up to episode $k$, and $(m_{ij}^\ell)_{\ell=1}^{M_k}$ are the realized service times of those completions.

\subsubsection{Time-varying arrivals}\label{sec:arrival_estimator}
\Cref{alg:learning_alg} takes as input the arrival rate of customers, i.e., the expected number of arrivals per time unit. 
The customer arrival rate in the call center is time-varying.
For example: the number of calls at night is lower than during the day time; during office hours there is usually a peak in call volume at the start of the work day; and on weekends and holidays the number of calls is less than on an average weekday. 
The arrival pattern for one week in shown in \Cref{fig:arrivals_May2001}. 
We observe that type-1 customers dominate the customer population, which is consistent throughout the entire data set. 
For the simulation input, we use the actual arrival times as reported in the data set. \\

{\bf Forecasting methods.}
To solve~\eqref{eq:LP_eps}, we use time-dependent estimators $\hat\lambda^k$ that are updated at the start of each episode $k=1,\dots$. 
Estimator $\hat\lambda_i^k$ forecasts the arrival rate of type-$i$ customers in the next episode. 
There are many different methods for time series forecasting \cite{Hyndman2021, Bastianin2019, Goldberg2014, Klein2024}. 
Some methods, including the Holt--Winters' seasonal method and (S)ARIMA and its extensions (e.g.\ SARIMAX), forecast the call volume for an entire day based on historical daily or seasonal patterns~\cite{Hyndman2021}.
For example, the SARIMA relies on autoregression, temporal differencing and moving averages and takes seasonality into account. 
However, these methods tend to rely heavily on the assumption that the future follows the same overall structure as the past, and they can perform poorly when the system exhibits abnormal behavior on otherwise regular (non-holiday) days.
Since (S)ARIMA models are typically fitted to longer-term daily or weekly cycles, they may not adapt quickly to unexpected fluctuations during the day, such as sudden surges in call volume due to external events, technical issues, or minor schedule changes.
To make our forecasting more robust to such variations, we instead forecast the arrival rate using only the data of the day itself. 
This allows the model to react faster to deviations from typical patterns while still accounting for underlying trends.
A wide range of such methods exist in literature, such as for example Generalized Additive Models and Bayesian Additive Regression Trees. For an overview, we refer to~\cite{Hyndman2021}.
However, in the spirit of explainability, and to reduce the risk of overfitting, we choose a simple forecasting method which we describe next. \\

{\bf Holt's linear trend model.}
We use Holt's linear trend model to forecast the call volume of episode $k$ using the realizations of previous episodes~\cite{Hyndman2021}.
The forecast $\hat{y}_i^{k+1}$ for the number of type-$i$ calls in episode $k+1$ is the sum of the level estimate $\ell_i^k$ and trend estimate $b_i^k$. 
The level estimate $\ell_i^k$ is the weighted average of the actual number of calls during episode $k$ and the forecast of the previous interval $k-1$, where the smoothing parameter $\alpha\in[0,1]$ controls the balance between reactivity to new data points and previous trends. 
Similarly, the trend estimate $b_i^k$ is the exponentially smoothed moving average of the trend $\ell_i^k - \ell_i^{k-1}$  with weight $\beta\in[0,1]$. 
The update equations are given by: 
\begin{align}
    \hat{y}_i^{k+1} &= \ell_i^k + b_i^k, \label{eq:holt_volume_est} \\
    \ell_i^k &= \alpha y_i^k + (1-\alpha)\hat{y}_i^{k-1}, \label{eq:holt_level_est} \\
    b_i^k &= \beta(\ell_i^k - \ell_i^{k-1}) + (1-\beta) b_i^{k-1}. \label{eq:holt_trend_est}
\end{align}
We initialize the forecast as $\hat{y}_i^1 := 0$ for all $i\in\calI$. 
The forecasted arrival rate $\hat{\lambda}_i^{k+1}$ for type-$i$ customers for episode $k+1$ is then the number of forecasted calls divided by the episode length $h$. 
Lastly, we take the maximum with zero since arrival rates cannot be negative: 
\begin{align}\label{eq:arr_est}
    \hat{\lambda}_i^{k+1} &= \max\Bigl(\frac{\hat{y}_i^{k+1}}{h}, 0\Bigr).
\end{align}

It remains to choose suitable values for $\alpha$ and $\beta$. 
To this end we analyzed the error between the forecast and the realizations of the training data for different values of $\alpha$ and $\beta$ and all customer types. 
Since the trend of the call volume differs per day and customer type, there is not one pair of values that outperforms all others. 
For $\alpha$, larger values lead to an overly reactive forecast that is sensitive to noise, while small values fail to pick up changes in the trend, which are especially present for type-1 customers.
For balance, we therefore set $\alpha = 0.5$ for the remainder of this paper. 
The larger $\beta$, the more sensitive the forecast is for rapid changes in the trend. 
However, most of the time, the trend changes only gradually. We therefore set $\beta=0.2$. \\

\Cref{fig:call_forecast} shows the true call volumes against the estimates given by~\eqref{eq:holt_volume_est}.
Since the trend estimate~\eqref{eq:holt_trend_est} is based on exponentially smoothed historical data, we observe a lag in the forecast when the trend in the data suddenly changes: 
for example, the call volume of type-1 customers is underestimated during the sudden increase between 6\AM and 9\AM, and overestimated when the volume drops between 9\AM and 11\AM. 
\Cref{fig:MASE}  analyzes the goodness--of--fit of Holt's linear trend model using the \gls{MASE} measure, which is defined as 
$
    1/n \sum_{k=1}^n |\hat{y}_i^k - y_i^k| / (1/(n-1) \sum_{\ell=2}^{n} |y_i^\ell - y_{i-1}^\ell|)
$,
where $n$ is the total number of intervals in a day. 
A \gls{MASE} value less than 1 indicates that the forecast outperforms the 1-point rolling average ($\hat{y}_i^k = y_i^{k-1}$).
In \Cref{fig:MASE}, we observe that Holt's model outperforms the 1-point rolling average on most days, and the 3-point rolling average ($\hat{y}_i^k = 1/3 \sum_{n=1}^3 y_i^{k-n}$) on all days. 
We conclude that Holt's linear trend model does a reasonable job forecasting the customer arrival rate, and that it outperforms naive methods on most days.

\begin{minipage}{.58\textwidth}
    \centering
    \begin{figure}[H]
        \centering
        \includegraphics[width=\textwidth]{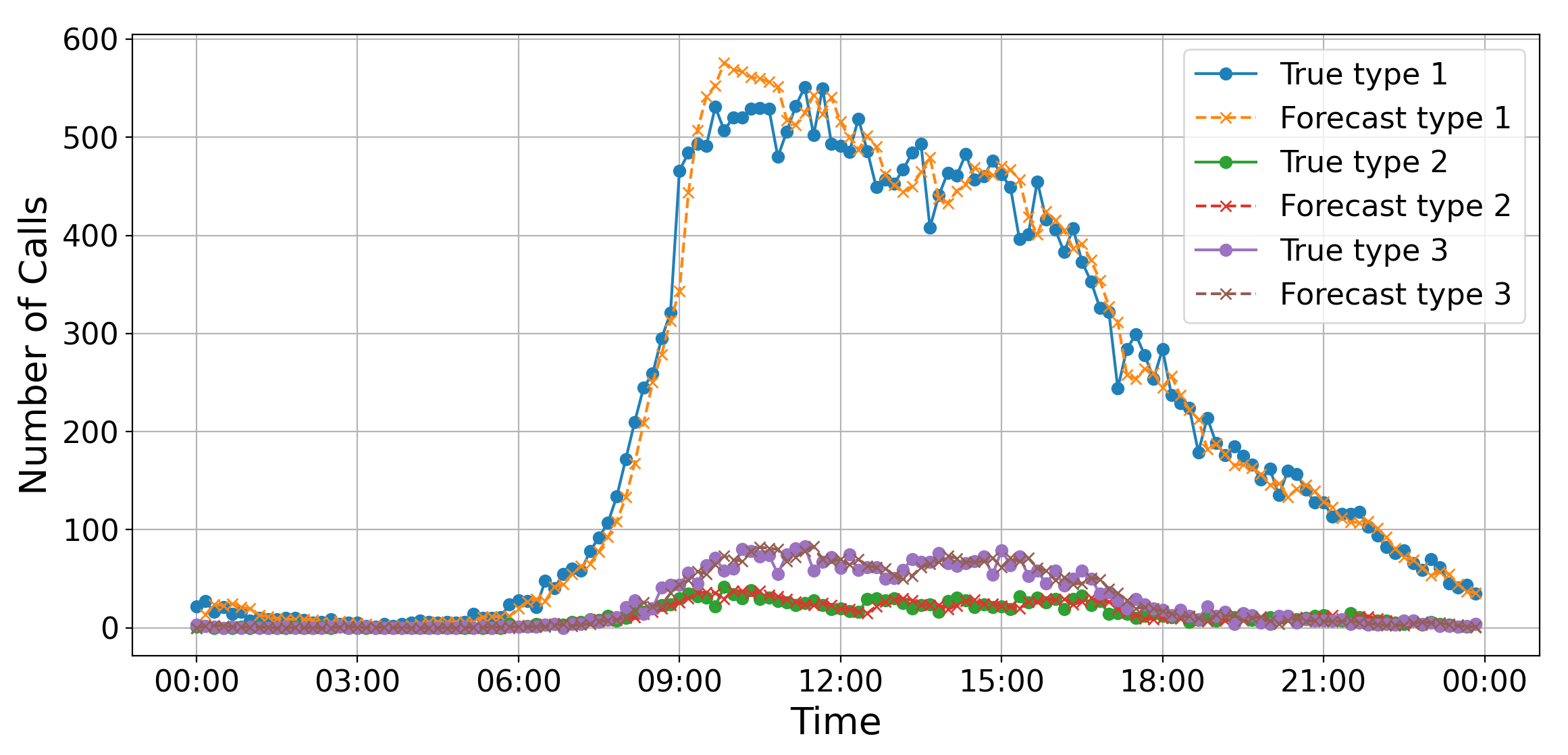}
        \caption{Forecast of the number of calls in 10-minute intervals for three different customer types on May 7, 2001 given by Holt's linear trend model~\eqref{eq:holt_volume_est} with parameters $\alpha = 0.5$ and $\beta = 0.2$.  }
        \label{fig:call_forecast}
    \end{figure}
\end{minipage}
\hspace{.5em}
\begin{minipage}{.4\textwidth}
    \centering
    \begin{figure}[H]
        \centering
        \includegraphics[width=\textwidth]{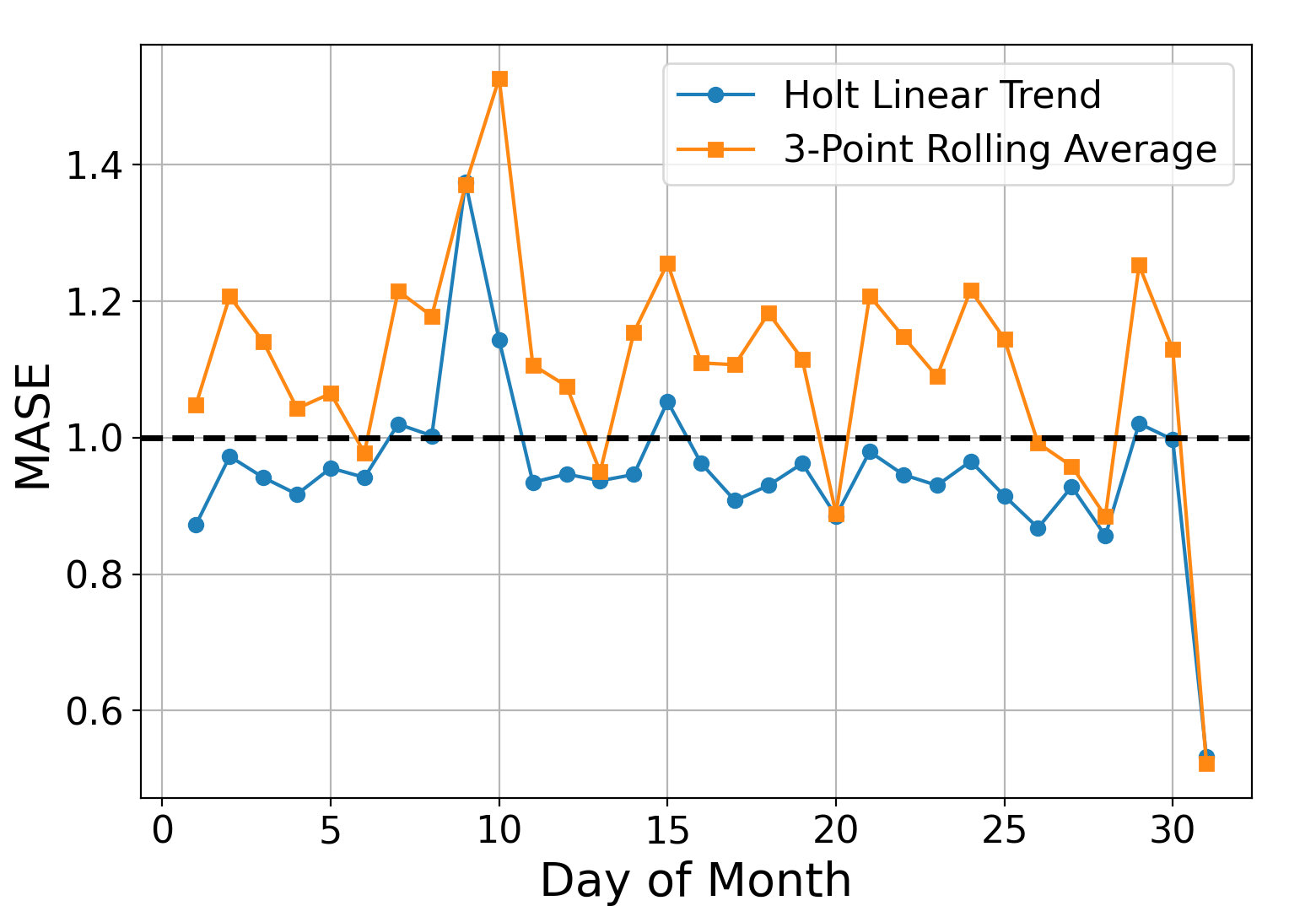}
        \caption{\gls{MASE} of Holt's linear trend model~\eqref{eq:holt_volume_est} and the 3-point rolling average for type-1 customers for May 2001.
        We note that May 9, 10, and 31 (Memorial Day) have an unusual arrival pattern.}
        \label{fig:MASE}
    \end{figure}
\end{minipage}

\section{Experiments}\label{sec:experiments}
In this section we conduct experiments using our simulation framework to demonstrate the potential of  \Cref{alg:learning_alg} (\gls{UCBQR}) in practical situations. 
In \Cref{sec:experiment_initial}, we analyze the performance of \gls{UCBQR} against commonly used benchmark routing policies. 
Since the results show significantly higher customer waiting times than the benchmarks, we introduce a new heuristic routing rule in \Cref{sec:alt_routing} to replace \Cref{alg:routing}.
With this, we show that the waiting times are successfully reduced and high payoffs are maintained.
In \Cref{sec:dif_obj}, we discuss  ways to adapt the \gls{UCBQR} algorithm to account for multiple objective functions and  show an example where payoff maximization and server fairness are jointly optimized.
\Cref{sec:est_error} analyzes the sensitivity against estimation errors in the arrival and service rates and \Cref{sec:episode_length} tests the sensitivity of the algorithm's performance to the episode length.
\\

To analyze the performance of \gls{UCBQR}, we consider  the following benchmark policies:
\begin{itemize}
    \item {\bf Oracle:} Similar to \Cref{alg:learning_alg}, but with oracle knowledge of the payoff, arrival and service time parameters. 
    In particular, this is \Cref{alg:learning_alg}, but with replacing $\hat{\theta}$, $\hat{\lambda}$, and $\hat{\mu}$ in \Cref{line:solve_LP_eps} and \Cref{line:solve_LP_adapted} by the actual $\theta$, $\lambda$, and $\mu$ as obtained from the data. 
    \item {\bf \gls{FCFS}---\gls{ALIS}:} If a server is idle and there are compatible customers waiting, assign the customer with the maximal waiting time. 
    On the other hand, if there are idle compatible servers upon a customer arrival, assign it to the server with maximal idle time. This classical routing policy is widely used in practice since it is easy to implement and both customers and servers experience a sense of fairness~\cite{Adan2014}. 
    \item  {\bf Greedy:} If server $j$ is idle and there are compatible customers waiting, assign the first--in--line customer with maximal expected payoff, i.e., a customer of type $\argmax_{i\in\calI} \theta_{ij}$. 
    If there are idle compatible servers upon a customer arrival, assign it to the server with maximal expected payoff $\theta_{ij}$. 
    This policy maximizes the instantaneous payoff without considering long-term effects or stability constraints. Note that it depends on the true payoff parameter $\theta$ and does not rely on  $\lambda$ and $\mu$. 
    \item {\bf Random:} If a server is idle and there are compatible customers waiting, choose a compatible queue uniformly at random and the first--in--line customer. 
    If there are idle compatible servers upon a customer arrival, assign the customer to a compatible idle server uniformly at random. 
    This policy serves as a most naive benchmark since it makes uninformed decisions. 
    \item {\bf $\theta\mu$ rule:} If server $j$ is idle and there are compatible customers waiting, assign the first--in--line customer with maximal $\theta_{ij}\mu_{ij}$. 
    If there are idle compatible servers upon a customer arrival, assign the customer to the server with maximal $\theta_{ij}\mu_{ij}$. 
    This policy is a variation of the classical $c\mu$ rule~\cite{Smith1956}, which is known to be optimal for holding cost minimization in a multi-customer single-server setting, and has been well-studied in multi-customer multi-server settings~\cite{Xia2022,Long2023} and even in adaptive learning settings~\cite{Krishnasamy2018b}.
    Instead of holding costs, we balance the average payoff $\theta_{ij}$ with service speed $\mu_{ij}$. 
    \item {\bf Data:}  We include the results from the data set which follow from the routing policy used by the call center. 
    It is unknown to us which policy was applied.
\end{itemize}

\subsection{Proof of concept}\label{sec:experiment_initial}
We apply \gls{UCBQR} on the data of June 2001. 
The pattern of arrivals, agent schedules, and the true value of $\theta$ are given in \ref{app:input_data_June2001}. 
We set the episode length to $h = 2$ minutes and we take $\eps = 10^{-6}$ and $p = 10^3$ in~\eqref{eq:LP_eps} and~\eqref{eq:LP_adapted}. 
Because the payoff parameters $\theta_{ij}$ lie in $[0,1]$ and the penalty $p$ is several orders of magnitude larger, giving weight to the variables $x_{iz}$, $i\in\calI$ is discouraged and only happens when there is no other feasible solution. 
For each day and policy, we run 50 independent simulations with the same arrivals from the data. \\

\begin{figure}[H]
    \centering
    \includegraphics[width=.9\linewidth]{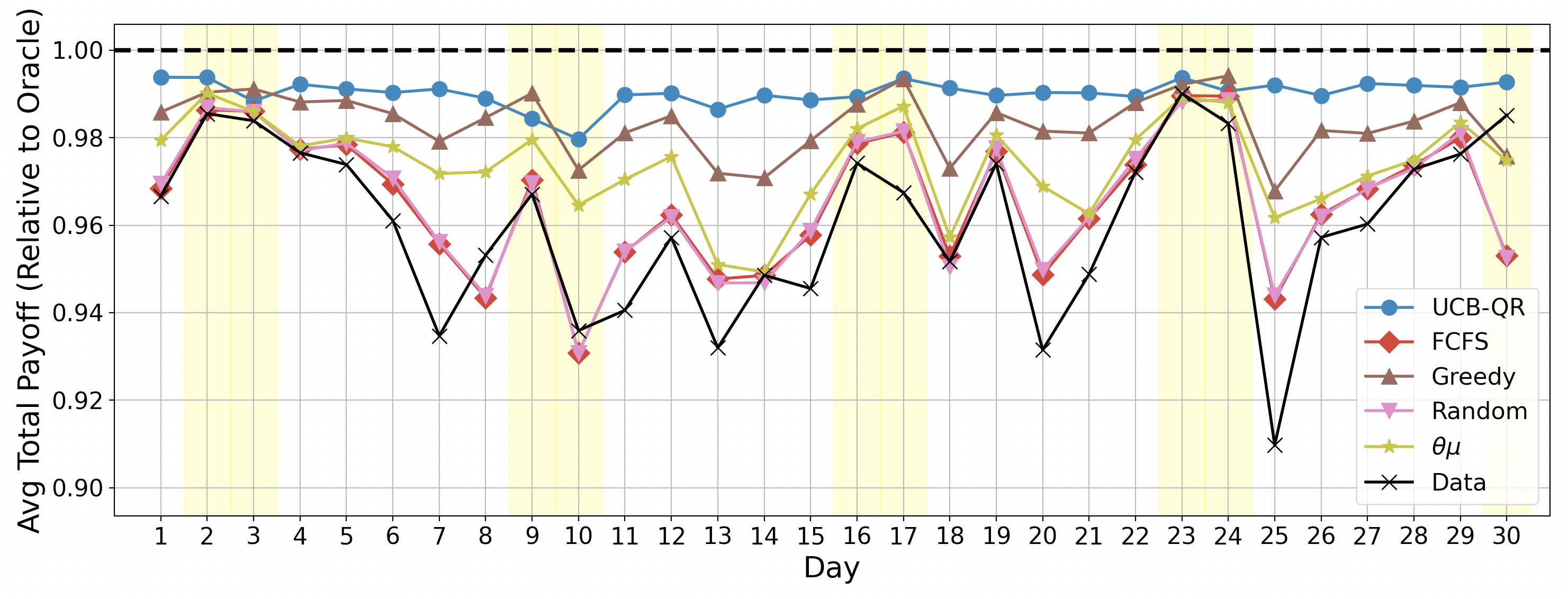}
    \caption{Average total payoff  per day in June 2001 expressed relative to the Oracle policy, for \Cref{alg:learning_alg} (\gls{UCBQR}), the benchmark policies, as well as the (unknown) routing policy used in the data set. 
    The variance from the average is negligible relative to the scale (10 compared to 50,000) and therefore omitted in the plot.  
    The days highlighted in yellow are weekends.}
    \label{fig:reward_June2001}
\end{figure}

\paragraph{\gls{UCBQR} exhibits little day--by--day fluctuations}
\Cref{fig:reward_June2001} shows the average total payoff per day obtained by the different policies. 
The values are scaled by the value of the Oracle policy, i.e., the plot shows $r_{\pi,i}/r_{O,i}$ where $r_{\pi,i}$ is the average payoff for policy $\pi$ on day $i$, and $O$ indicates the Oracle policy. 
The call volume is different for each day, as shown in \ref{app:input_data_June2001}, with fewer calls on weekends. 
The payoffs reflect the quality of the customer--server matches; higher values indicate better performance.
Observe that the \gls{UCBQR} algorithm performs closely to the Oracle across all days.
The loss in payoff due to learning $\theta$ and estimating $\lambda$ and $\mu$ is  $0.7-1\%$ on week days and $1-2\%$ during the weekend. 
The \gls{FCFS}---\gls{ALIS}, Random, $\theta\mu$ policies and the policy from the data set obtain lower payoffs and show strong fluctuations over the month in contrast to the more stable performance of \gls{UCBQR}. 
Especially on June 25, most policies show a decrease in payoff except for \gls{UCBQR}.
This showcases the potential of the algorithm to learn and adapt under different load patterns. \\

\paragraph{Greedy sometimes performs best on weekends, but \gls{UCBQR} consistently reduces the optimality gap}
Note that the differences in payoff accumulation in \Cref{fig:reward_June2001} between the different policies are relatively small. 
For weekend days 3, 9, and 24 June, the Greedy policy, which is designed for myopic payoff maximization, achieves a higher payoff than \gls{UCBQR}.
This can be explained by the value of the payoff parameter $\theta$ 
(see e.g.\ \Cref{tab:true_theta_4June2001} in \ref{app:input_data_June2001}): 
the vector shows little variation over the different customer types and agent groups, especially for type-1 customers, while these customers dominate the call volume.
We observe that in this case the Random policy performs quite well in terms of payoff and accumulates on average 96.5\% of the Oracle payoff. 
The room for improvement is therefore small in this example: the optimality gap that measures the loss in payoff with respect to the Oracle payoff is just 3.5\%. 
However, observe that the \gls{UCBQR} policy obtains on average 99.0\% of the Oracle payoff, closing the optimality gap of the Random policy by approximately $71.4\%$. \\

\paragraph{\gls{UCBQR} outperforms benchmarks under high variance payoffs}
To gain a better insight in the payoff accumulation of \gls{UCBQR}, we conduct the same experiment using a transformed version of $\theta$ that shows more variance over the different customer--agent combinations in \ref{app:adapted_theta}.
The payoff accumulation in this case shows a clear benefit of \gls{UCBQR} over the benchmark policies and the policy of the data set: for all days, the \gls{UCBQR} policy achieves at least $98\%$ of the Oracle payoff while other policies obtain as little as $60\%$ on some days. 
Overall, the ordering of the policies is the same as in \Cref{fig:reward_June2001}, i.e., the Greedy and $\theta\mu$ policies perform similarly and better than the \gls{FCFS}---\gls{ALIS}, Random, and Data policies. 
To conclude, \gls{UCBQR} achieves overall the best payoff accumulation, and the relative difference in payoff gain depends on $\theta$. 

\paragraph{\gls{UCBQR} does not minimize waiting times}
\Cref{fig:wait_4June2001} shows the waiting times of customers on June 4, and \Cref{fig:boxplot_wait_4June2001} shows the average waiting times between 6\AM and 9\PM (office hours).
In the data set, there is a sharp peak in waiting time around midnight, while all other policies maintain more evenly distributed waiting times during the night.
Around 6\AM, the offered load increases and many agents start their work.
Between 6\PM and 9\PM the policy from the data set shows another peak, which is handled more smoothly by the Oracle policy and \gls{UCBQR}, indicating a better way of adapting to a change in load. 
The benchmark policies \gls{FCFS}---\gls{ALIS}, Greedy, Random, and $\theta\mu$ consistently give low waiting times, which is why the plots are at some places indistinguishable. 
Overall, the waiting times for the \gls{UCBQR} and the Oracle policy are significantly higher than those of the benchmark policies and the policy used in the data set. 

\begin{minipage}{.66\textwidth}
    \centering
    \begin{figure}[H]
        \centering
        \includegraphics[width=1\textwidth]{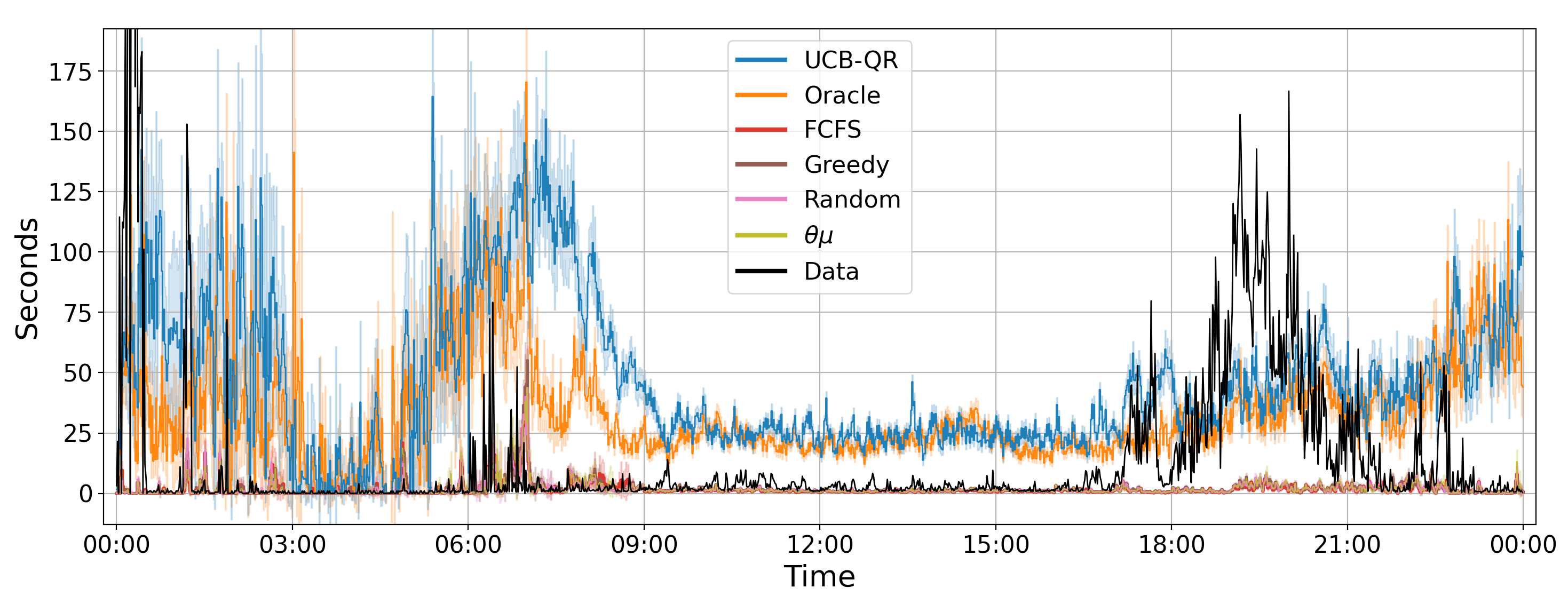}
        \caption{Average waiting times of customers over all classes with 95\% confidence intervals, binned in 60 second intervals when applied to the data of June 4, 2001.}
        \label{fig:wait_4June2001}
    \end{figure}
\end{minipage}
\hspace{.5em}
\begin{minipage}{.3\textwidth}
    \centering
    \begin{figure}[H]
        \centering
        \includegraphics[width=\textwidth]{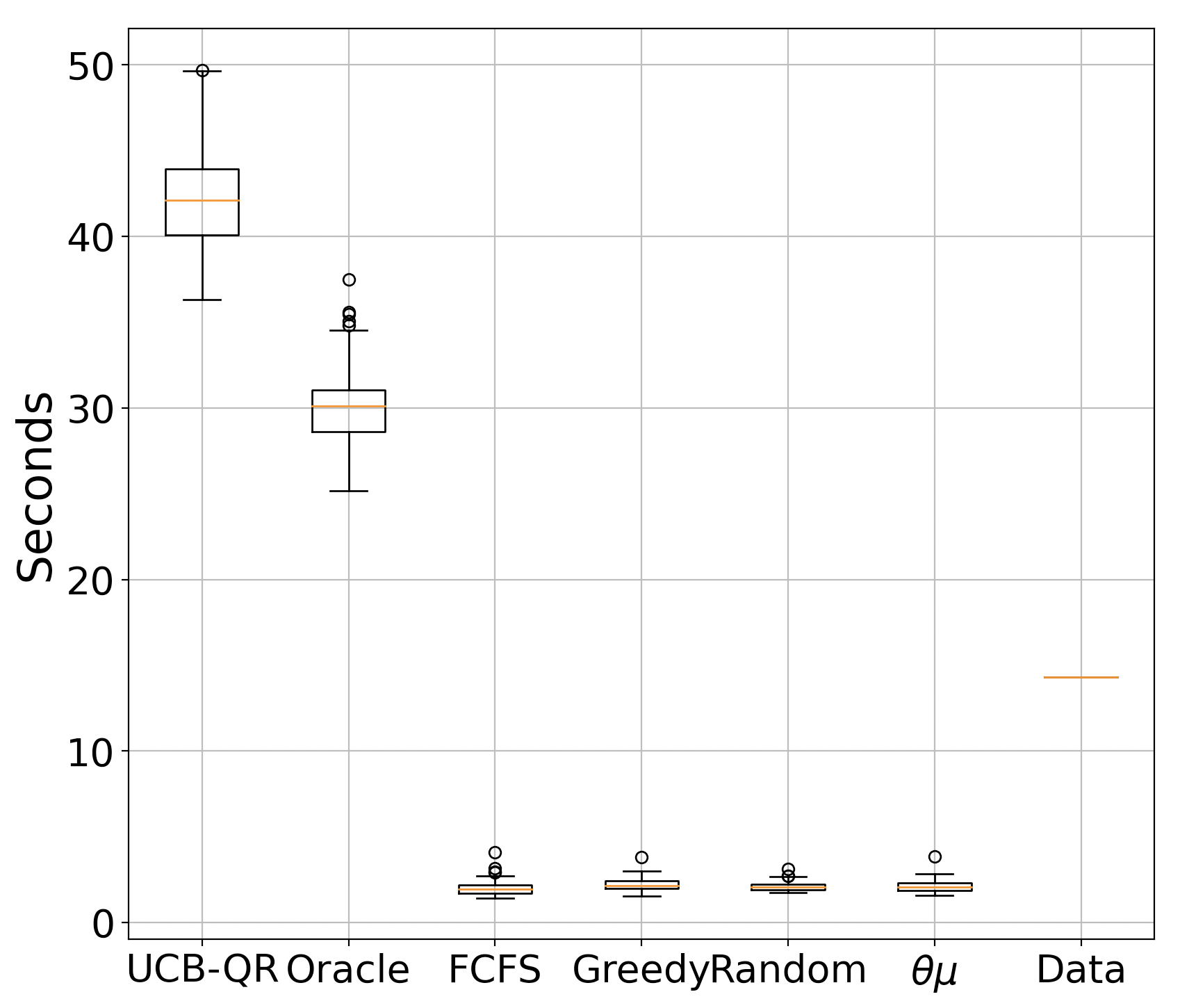}
        \caption{Boxplot of waiting time during office hours (6\AM-9\PM) when applied to the data of June 4, 2001.}
        \label{fig:boxplot_wait_4June2001}
    \end{figure}
\end{minipage}

\begin{figure}[H]
    \centering
    \includegraphics[width=\linewidth]{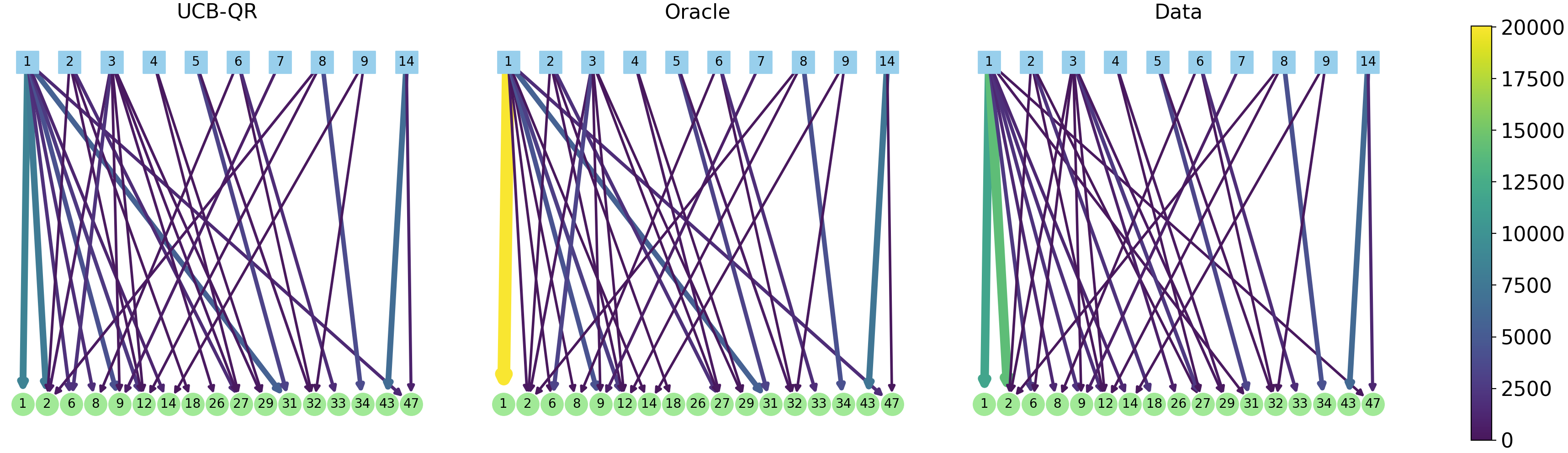}
    \caption{Routing volume for various routing policies when applied to the data of June 4, 2001.}
    \label{fig:routing_4June2001}
\end{figure}

\paragraph{Routing behavior of \gls{UCBQR} is affected by exploration}
\Cref{fig:routing_4June2001} illustrates the total routing volume of customers to the different servers for \gls{UCBQR} and the Oracle policy, as well as the policy from the data set. 
\gls{UCBQR} routes customers similar to the Oracle policy, but deviations are due to learning the payoff parameter and estimating the arrival and service rates. 
The Oracle policy does not utilize server 26 at all, since the average payoffs for this server are lower than those at other servers (see \ref{app:input_data_June2001}), and stability can be maintained without the capacity of this server. 
The \gls{UCBQR} policy does use server 26 since it needs to explore the different compatible lines. 
The Oracle policy routes type-1 customers mainly to server 1, while \gls{UCBQR} and the policy from the data set balances the load between servers 1 and 2. 
Server loads are further discussed in \Cref{sec:dif_obj}.
Overall, we observe that exploration of the \gls{UCBQR} policy leads to different routing decisions when compared to the Oracle policy, although the overall payoff accumulation is competitive. 

\paragraph{Computational complexity of \gls{UCBQR} matches benchmarks}
All simulations were performed on separate threads using an Apple M1 Pro-chip with 8-core CPU.
Running a full-day simulation of the \gls{UCBQR} algorithm takes about 20 seconds.
Although this is twice as long as a typical runtime of roughly 10 seconds for the benchmark policies, it is still insignificant given that the simulation covers a 24-hour period.

\subsection{Alternative routing rule to decrease waiting times}\label{sec:alt_routing}
We now take a critical look at the routing rule \Cref{alg:routing} used by \gls{UCBQR} and the Oracle  policy.
We note that \Cref{alg:routing} is not very adaptive, since customers are assigned to a server (via the server's virtual queue) upon arrival and are not allowed to be routed to another idling server even when the assigned server has a long virtual queue. 
\Cref{alg:routing} also requires virtual queues which increases the complexity of the simulation, and the reshuffling of the virtual queues between episodes (\Cref{line:reallocation2}) can be experienced as unfair from a viewpoint of both customers and servers in practice, since service is not \gls{FCFS} (not even for customers of the same type) nor \gls{ALIS}. 
We may therefore try to construct an improved routing rule with less delays, while still routing according to the desired rates described by the \gls{LP}. 
Concretely, the empirical routing rates $\E[D_{ij}(t)]/t$, where $D_{ij}(t)$ is the number of type-$(ij)$ departures up to time $t$, should converge to the desired rate $x_{ij}$.
To the best of our knowledge, not many such rules exist in literature, since routing rates and even stability conditions are generally difficult to determine for arbitrary routing rules for skill-based queues~\cite{Adan2012,Weiss2021}. \\

Yet, it is more feasible to heuristically justify a routing policy that allocates customers in a matter that approximates the desired routing rates. 
To this end, we propose and test \Cref{alg:tree}, which takes inspiration from the heuristic \gls{JSQK} policy used in~\cite{Fu2022}  (not to be confused with the \gls{JSQK} algorithm of~\cite{Ephremides1980}).
\Cref{alg:tree}  takes as input routing rates $x\in\R^{|\calL|}$, the maximizer of the \gls{LP} \eqref{eq:LP_eps} or \eqref{eq:LP_adapted} in case the former is infeasible, a payoff vector $\theta\in\R^{|\calL|}$, and the episode length $h$. 
We stress that $\hat{\theta}$ is used instead of the true value $\theta$ in \Cref{alg:learning_alg}. 
Each customer remains in the queue of its own type until it is assigned to an idle server, making virtual queues obsolete. 
Since the routing rate vector $x$ is a \gls{BFS} of an \gls{LP}, it induces a spanning forest on the bipartite graph of customer types, servers, and compatibility lines $(\calI\cup\calJ,\calL)$ \cite[Proposition 2]{Fu2022}, \cite{Bertsimas1997}. 
That is, the queues (consisting of customers of its own type) and servers are the nodes of the spanning forest and the edges are the queue--server connections with non-zero weight in the solution $x$. 
In such a forest, the parent and child nodes of servers are queues and \emph{vice versa}. 
The root of the tree is a slack server, which has only child queues and no parent queue. This is possible since every tree in the spanning forest contains at least one slack server satisfying $\sum_{(ij)\in\calL} x_{ij}/\mu_{ij} < 1-\eps$, and exactly one slack server if $x$ is non-degenerate~\cite[Proposition 2]{Fu2022}. Upon service completion, the server prioritizes customers of their child queue(s), and only if all child queues are empty, the server serves customers from its parent queue (if any). This way, non-slack servers are highly utilized, and the priority rule encourages routing decisions according to the target routing rates $x$ corresponding to the spanning forest.

\begin{algorithm}[H]
    \caption{Tree-based routing($x, \theta, h$).}
    \label{alg:tree}
    \begin{algorithmic}[1]
        \State Consider the spanning forest of the bipartite graph $(\calI\cup\calJ,\calL)$ induced by $x$. 
        \For{time $t = 0,\dots, h$} \Comment{Route customers.} 
            \For{each arrival of a type-$i$ customer at time $t$}
                \If{queue $i$ has idling children servers in the spanning forest}
                    \State Start service at the idle child server $j$ that maximizes $\theta_{ij}$. 
                \ElsIf{the parent server of queue $i$ is idle}
                    \State Start service at the parent server. 
                \Else
                    \State Customer joins queue $i$. 
                \EndIf
            \EndFor
            \For{each service completion at server $j$ at time $t$}
                \If{server $j$ has non-empty children queues}
                    \State Start service of the first--in--line customer of the non-empty child queue $i$ that maximizes $\theta_{ij}$. 
                \ElsIf{server $j$ has a non-empty parent queue $i$}
                    \State Start service of the first--in--line customer of the parent queue. 
                \Else
                    \State Server $j$ remains idle. 
                \EndIf
            \EndFor
        \EndFor
    \end{algorithmic}
\end{algorithm}

\begin{figure}[H]
    \centering
    \includegraphics[width=.9\linewidth]{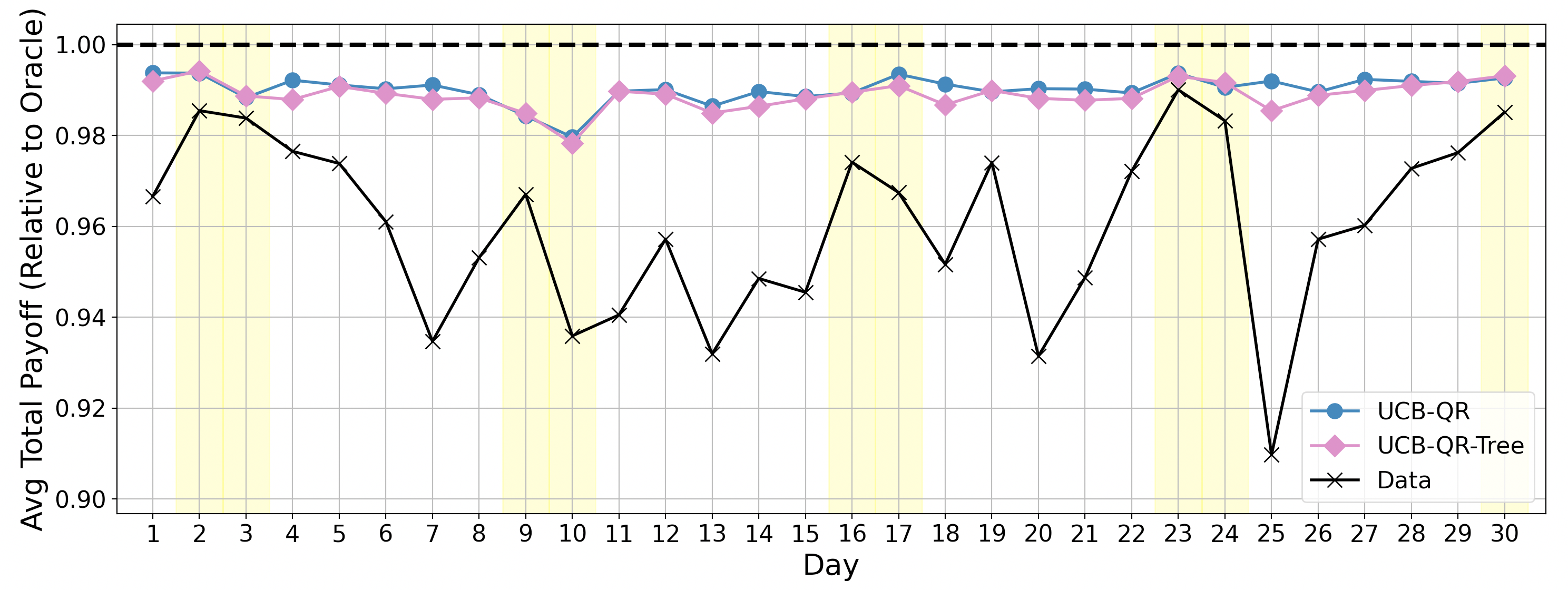}
    \caption{Average total payoff per day in June 2001 for the \gls{UCBQR} and \gls{UCBQR}--Tree policies. }
    \label{fig:reward_tree_June2001}
\end{figure}

\paragraph{Minimal payoff loss from heuristic routing rule}
We compare the performance \Cref{alg:learning_alg} with an adapted version, called \gls{UCBQR}--Tree, which uses \Cref{alg:tree} as routing rule instead of \Cref{alg:routing}.
We found that the empirical routing volumes of customers to servers for  \gls{UCBQR}--Tree were similar to those of \gls{UCBQR}, implying that the accumulated payoff is minimally effected as shown in \Cref{fig:reward_tree_June2001}.
By construction, we can intuitively argue that within each episode the empirical routing rates of \Cref{alg:tree} should closely approximate their target $x$. 
Yet, from a small example in \ref{app:routing_q_small} it can be seen that the empirical routing rates $\E[D_{ij}(t)]/t$ of \Cref{alg:tree} do not converge to the target $x_{ij}$ exactly, but only to values close to it.
A theoretical analysis of the empirical routing rates remains a direction for future work. 
For a practical implementation, however, such minor deviations have little impact on performance measures including total payoff and waiting time, especially since there is inherent noise in the arrival and service rate estimators. 

\begin{figure}[H]
    \centering
    \hspace{-1em}
    \begin{subfigure}{0.47\linewidth}
        \centering
        \includegraphics[width=1.05\textwidth]{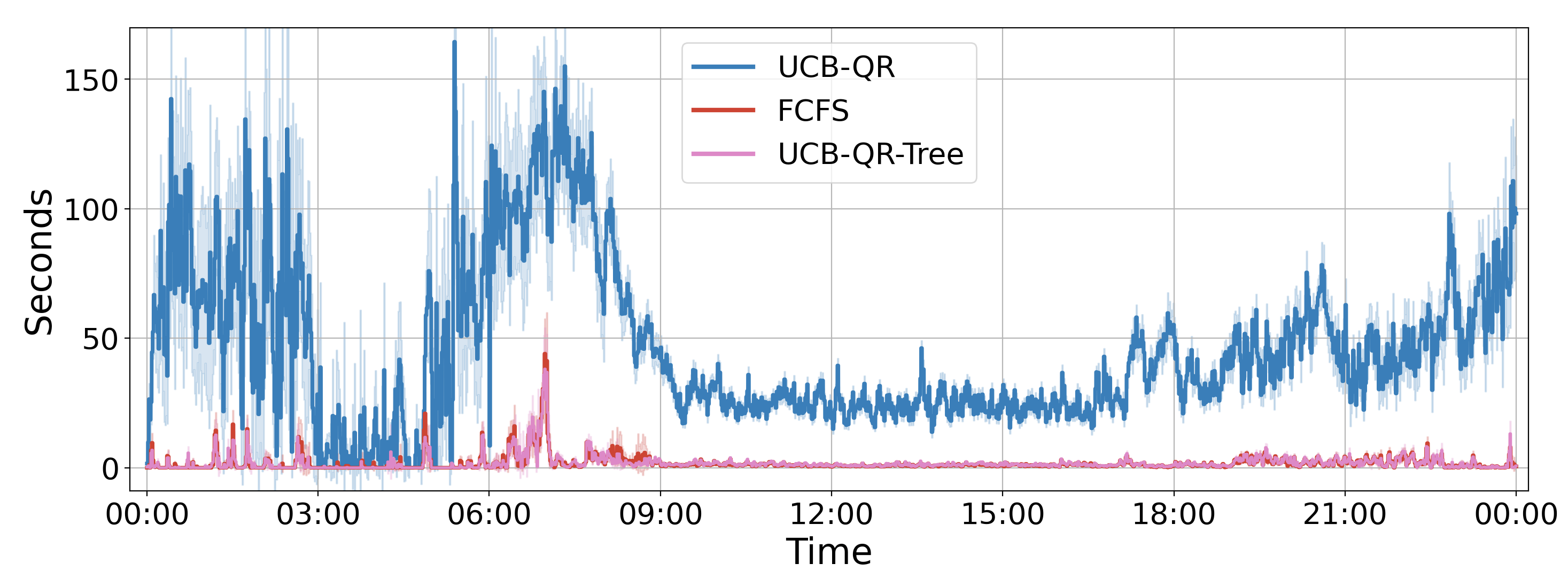}
        \caption{June 4, 2001.}
        \label{fig:wait_tree_4June2001}
    \end{subfigure}
    \hspace{0.01\linewidth}
    \begin{subfigure}{0.47\linewidth}
        \centering
        \includegraphics[width=1.05\textwidth]{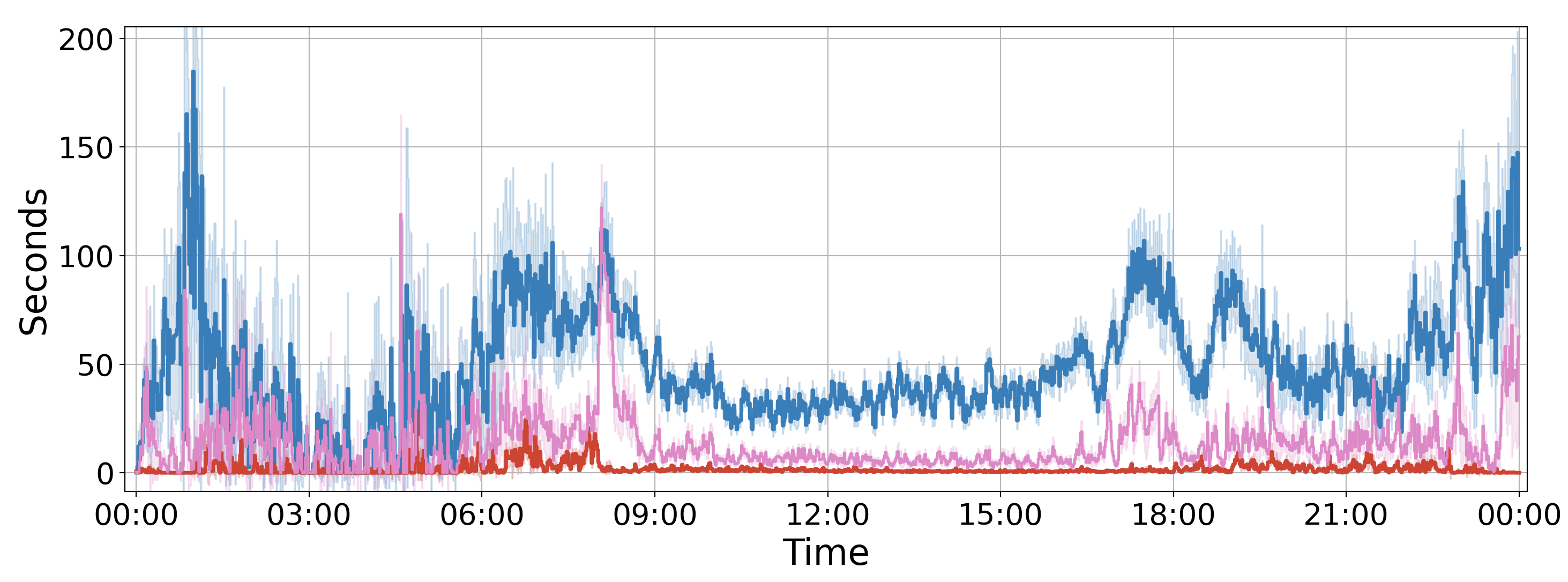}
        \caption{June 5, 2001.}
        \label{fig:wait_tree_5June2001}
    \end{subfigure}
    \caption{Average customer waiting time for the \gls{UCBQR} and \gls{UCBQR}--Tree policies. }
    \label{fig:wait_tree}
\end{figure}

\paragraph{\gls{UCBQR}--Tree significantly reduces waiting times}
The waiting times for \gls{UCBQR}--Tree are lower than those of \gls{UCBQR}, as shown in \Cref{fig:wait_tree}. 
In \Cref{fig:wait_tree_4June2001}, \gls{UCBQR}--Tree achieves waiting times almost as low as those of \gls{FCFS}---\gls{ALIS}. 
On the depicted day, the load is relatively low since there are many agents at work while the total call volume is moderate for a Monday (see \Cref{fig:arrivals_4June2001} and \Cref{fig:agent_schedule_4June2001} in \ref{app:input_data_June2001}). 
However, on days with higher load the average delay of \gls{UCBQR}--Tree is slightly worse than \gls{FCFS}---\gls{ALIS}, although it still outperforms \gls{UCBQR} as shown in \Cref{fig:wait_tree_5June2001}. 

\paragraph{\gls{UCBQR} and \gls{UCBQR}--Tree have comparable runtimes}
The computational complexity of \gls{UCBQR}--Tree is lower than \gls{UCBQR}, since virtual queues and reshuffling are omitted. 
However, at the start of each episode, the spanning forest must be determined based on the \gls{LP} maximizer. 
The runtimes were not substantially affected in our specific experiments. \\

In conclusion, \Cref{alg:tree} is an attractive alternative to \Cref{alg:routing} for practical applications, where exact convergence to target routing rates is not crucial while low waiting times are. 
The \gls{UCBQR}--Tree policy still accumulates a similar high payoff as \gls{UCBQR} while waiting times are reduced significantly. 
It additionally promotes customer fairness in the sense of \gls{FCFS} service for customers of the same type.

\subsection{Combining different performance objectives}\label{sec:dif_obj}
The \gls{UCBQR} algorithm is designed to maximize the total payoff, which is reflected in the objective function of \eqref{eq:LP_eps}.
In practice, however, service system management typically accounts for multiple performance criteria simultaneously.  
Next, we demonstrate how the \gls{UCBQR} algorithm can be extended to cover other criteria as well, as long as they can be expressed in terms of routing rates $x_{ij}$, $(ij)\in\calL$. We first present an illustrative example that balances payoff maximization and server fairness, followed up by a discussion on how to achieve such integration for other objectives as well.\\

To balance payoff maximization and server fairness, we adapt the objective function of the LP~\eqref{eq:LP_eps} to account for fairness, see optimization problem~\eqref{eq:OPT_eps}. 
Here, $\rho_j := \sum_{i\in\calI} x_{ij}/\mu_{ij}$ represents the load of server $j$, and the penalty terms $\rho_j - \bar{\rho}$ penalize deviation from the mean $\bar{\rho} := 1/|\calJ| \sum_{j\in\calJ} \rho_j$.
The parameter $\gamma\in\R$ controls the balance between server fairness and payoff maximization: 
the larger $\gamma$, the more weight is given to server fairness, at a cost of loss in payoff. 
Note that \Cref{alg:learning_alg} can simply be applied with this alternative optimization problem~\eqref{eq:OPT_eps} replacing \gls{LP}~\eqref{eq:LP_eps}.
For our simulation results next, we used the convex solver MOSEK~\cite{MOSEK2023}. 

\begin{subequations}
    \begin{align}
        \textrm{OPT}(\lambda, \mu, \theta,\eps,\gamma): \ \ \max_x \ \ &\sum_{(ij)\in\calL} \theta_{ij} x_{ij} - \gamma \sum_{j\in\calJ} (\rho_j - \bar{\rho})^2, \\
        \textrm{s.t.} \ \ &\sum_{j\in\calS_i} x_{ij} = \lambda_i, \ \ \ \forall i\in\calI, \\
        &\sum_{i\in\calC_j} \frac{x_{ij}}{\mu_{ij}} \leq 1 - \eps, \ \ \ \forall j\in\calJ,  \\
        &x_{ij} \geq 0, \ \ \ \forall (ij)\in\calL. 
    \end{align}
    \label{eq:OPT_eps}
\end{subequations}

\begin{figure}[H]
    \centering
    \hspace{-1em}
    \begin{subfigure}{0.48\linewidth}
        \centering
        \includegraphics[width=\textwidth]{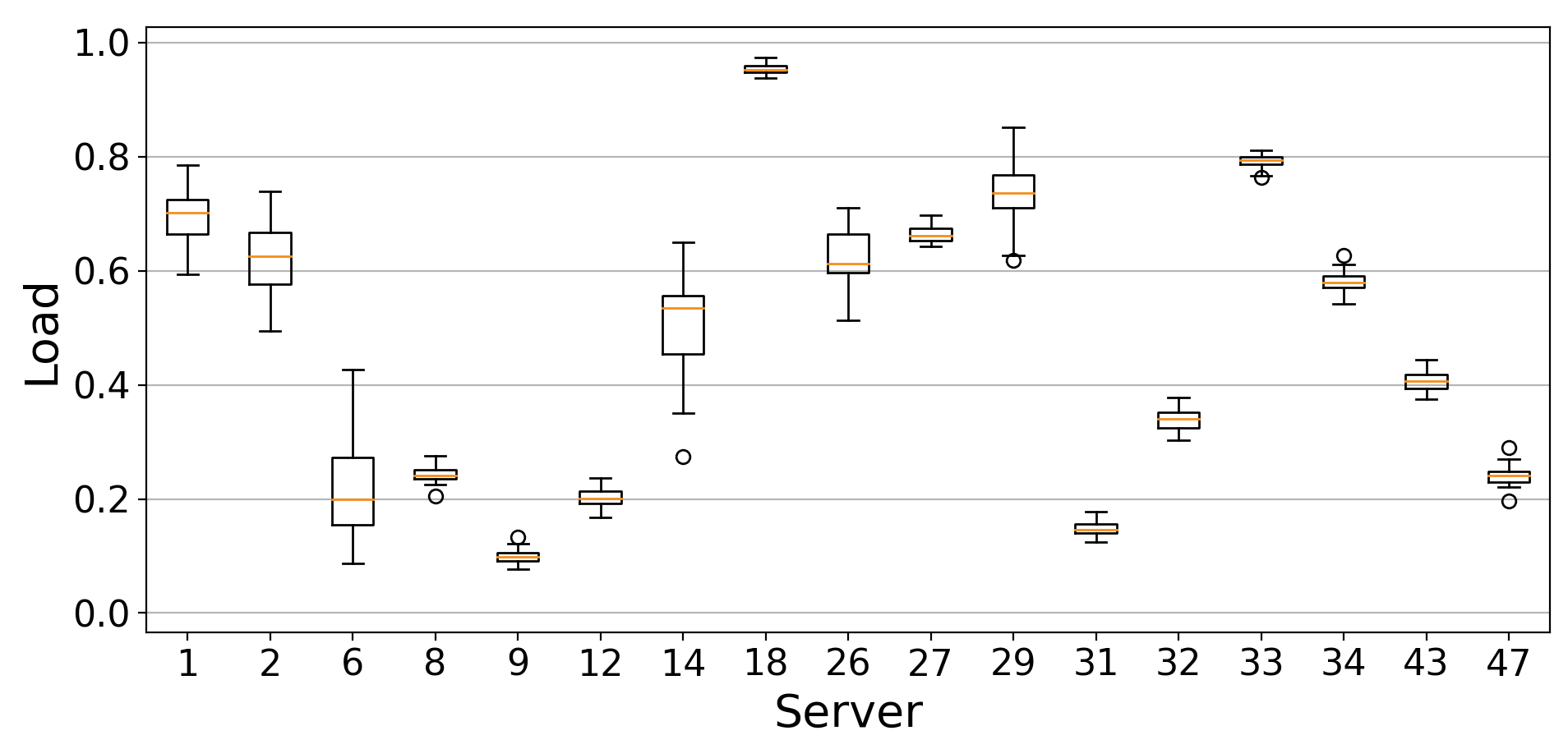}
        \caption{$\gamma=0$.}
        \label{fig:load_gamma_0}
    \end{subfigure}
    \hspace{0.01\linewidth}
    \begin{subfigure}{0.48\linewidth}
        \centering
        \includegraphics[width=\textwidth]{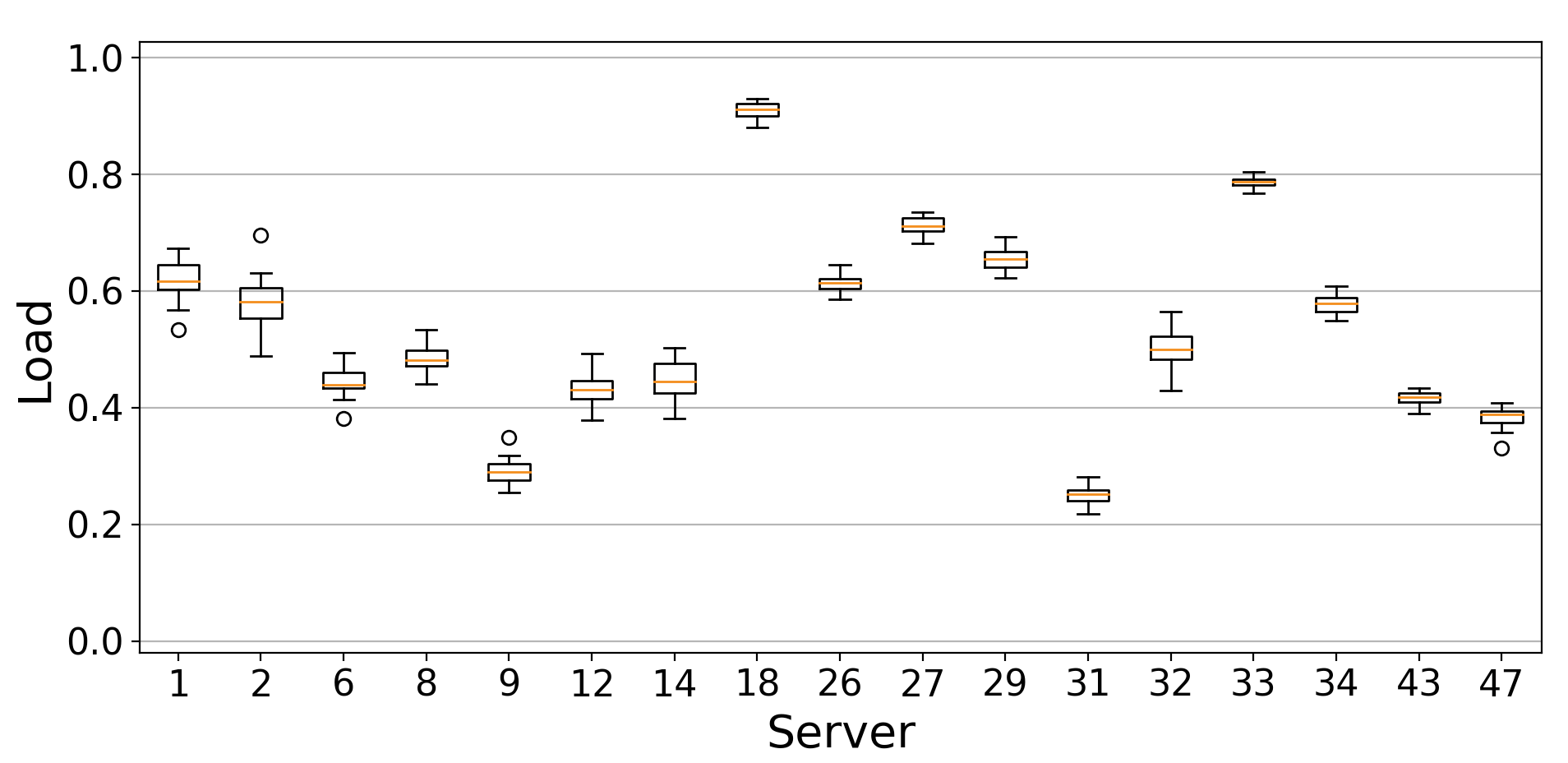}
        \caption{$\gamma=0.01$.}
        \label{fig:load_gamma_001}
    \end{subfigure}
    \begin{subfigure}{0.48\linewidth}
        \centering
        \includegraphics[width=\textwidth]{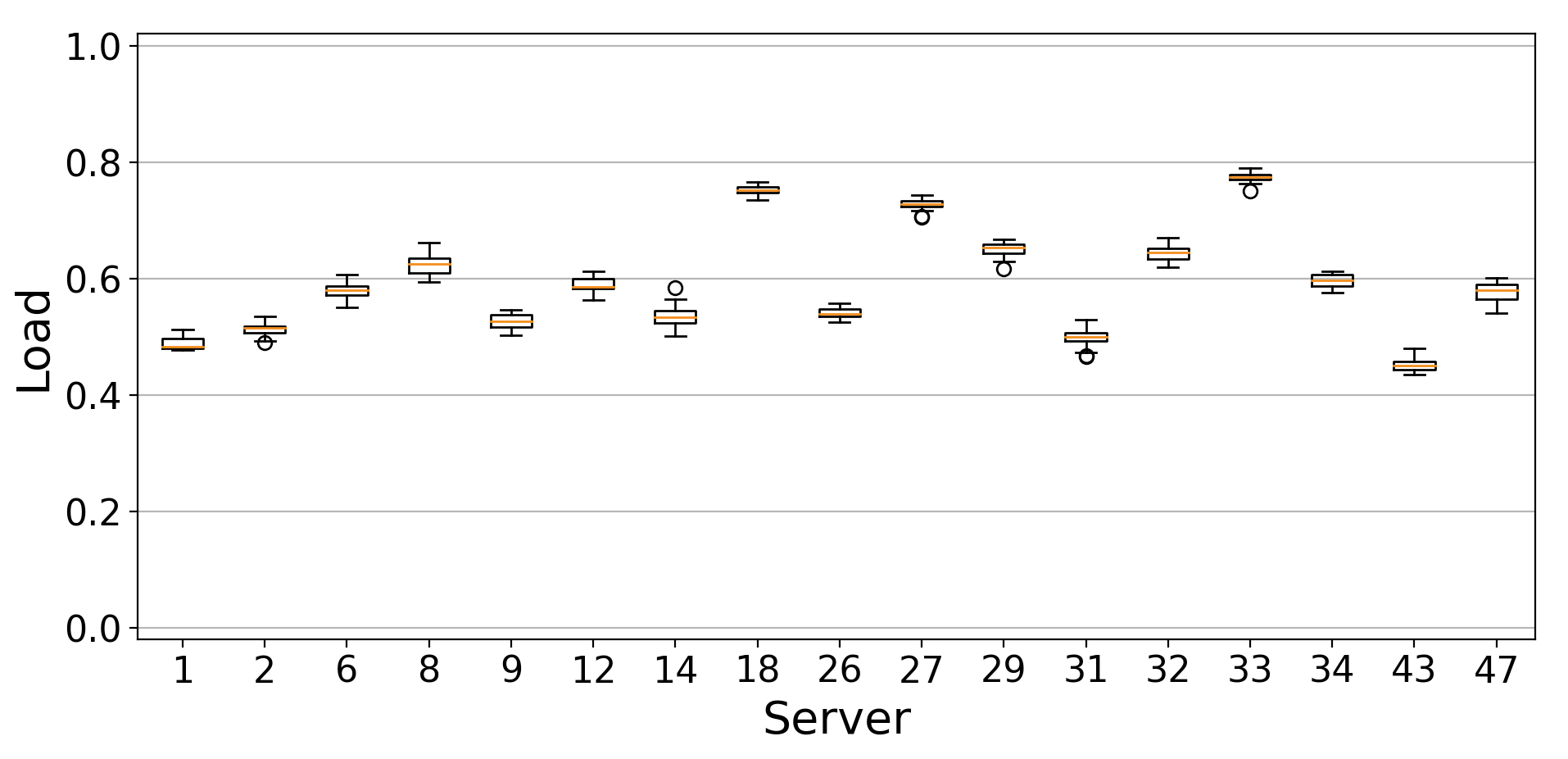}
        \caption{$\gamma=0.1$.}
        \label{fig:load_gamma_01}
    \end{subfigure}
    \begin{subfigure}{0.48\linewidth}
        \centering
        \includegraphics[width=\textwidth]{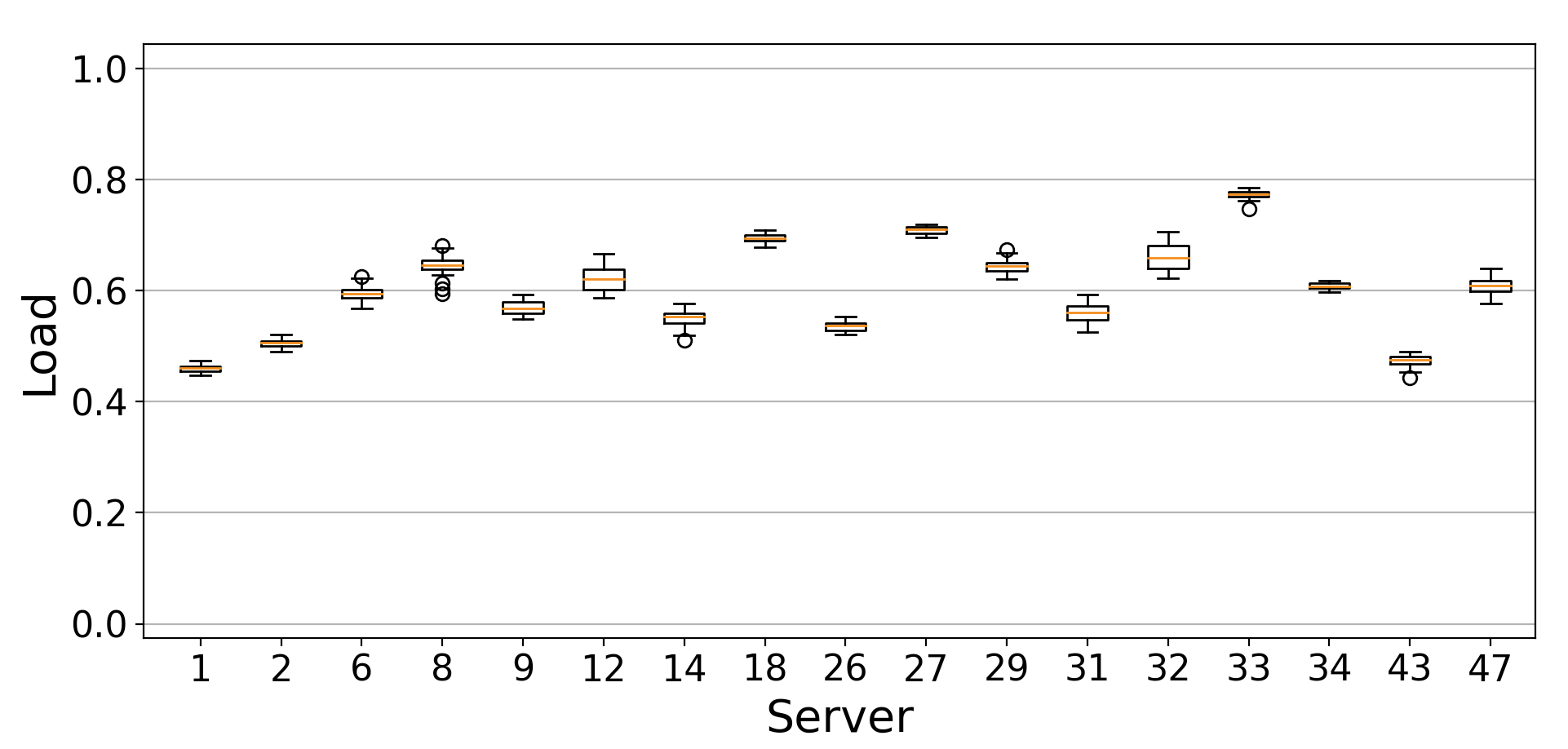}
        \caption{$\gamma=1$.}
        \label{fig:load_gamma_1}
    \end{subfigure}
    \caption{Empirical server load during office hours for \gls{UCBQR} with objective function~\eqref{eq:OPT_eps} for different values of $\gamma$ when applied to the data of June 4, 2001.}
    \label{fig:load_gamma}
\end{figure}

\begin{figure}[H]
    \centering
    \hspace{-1em}
    \begin{subfigure}{0.69\linewidth}
        \centering
        \includegraphics[width=\textwidth]{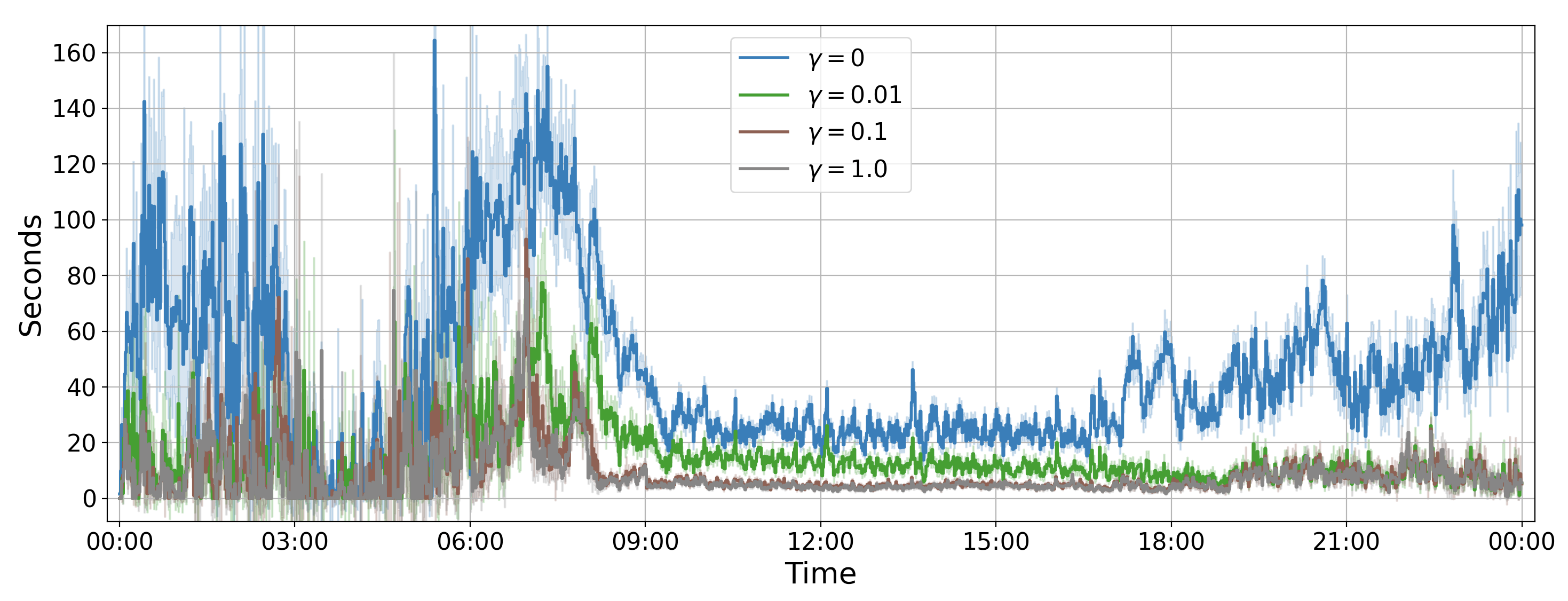}
    \end{subfigure}
    \hspace{0.01\linewidth}
    \begin{subfigure}{0.2\linewidth}
        \centering
        \includegraphics[width=1.05\textwidth]{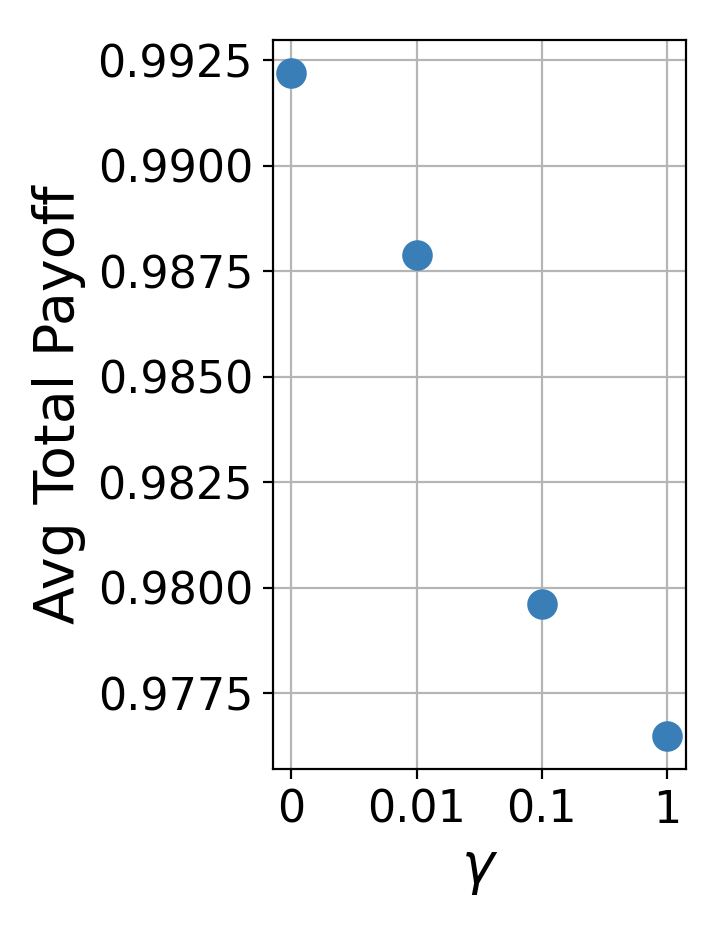}
    \end{subfigure}
    \caption{Average customer waiting time and average payoff (relative to the Oracle policy) of \gls{UCBQR} with objective function~\eqref{eq:OPT_eps} for different values of $\gamma$ when applied to the data of June 4, 2001.}
    \label{fig:wait_payoff_gamma_4June2001}
\end{figure}

\paragraph{Load balancing decreases waiting times with minimal payoff loss}
We observe in \Cref{fig:load_gamma} that the server loads indeed become more concentrated as the value of $\gamma$ increases. 
As a result, the variance in the load per server also decreases. 
Here, $\gamma=0$ represents the original \gls{LP} \eqref{eq:LP_eps}. 
This redistribution of work also affects customer waiting times:
when more servers are actively involved in processing customers the system is less likely to develop long queues at critically loaded servers, which in turn reduces the overall waiting time as $\gamma$ increases.
However, because customers are not always matched to their highest payoff server, the total accumulated payoff becomes slightly lower.
This trade--off is illustrated in \Cref{fig:wait_payoff_gamma_4June2001}, which shows how increasing $\gamma$ leads to shorter waiting times with only a modest decline in total payoff.

\paragraph{Incorporating other objectives}
A similar approach can be used to integrate other objectives: we formulate a convex objective for the performance measure in terms of routing rates and add it with a trade--off parameter to the objective function in \eqref{eq:LP_adapted}. 
Some examples are given in \Cref{tab:objectives}. 
Alternatively, bounds on the performance measures can be imposed by formulating new constraints for~\eqref{eq:LP_adapted}.
For instance, server loads can be bounded using $\underline{\delta}_j \leq \rho_j \leq \overline{\delta}_j$ for $\underline{\delta}_j, \bar{\delta}_j \in[0,1]$.

\begin{table}[H]
    \centering
    \footnotesize
    \begin{tabular}{cccc}
    \toprule
    Penalize critical server load & Penalize high waiting time & Penalize deviation from $x^0$ & Encourage spare routing \\
    \midrule
    $\sum_{j\in\calJ} \frac{1}{1-\rho_j}$ & $\sum_{j\in\calJ}\frac{\rho_j}{1-\rho_j}$ & $\sum_{(ij)\in\calL} (x_{ij} - x_{ij}^0)^2$ & $\sum_{(ij)\in\calL}\log(1+x_{ij})$ \\
    \bottomrule
    \end{tabular}
    \caption{Some examples of performance objectives expressed in terms of routing rates $x_{ij}$, $(ij)\in\calL$. }
    \label{tab:objectives}
\end{table}

\paragraph{Heuristic routing rule is incompatible with non-linear objectives}
We note that routing rule \Cref{alg:tree} is not implementable in its current form with a general optimization problem, since it relies on the properties of \glspl{LP}.
Specifically, it uses the spanning forest on the bipartite graph of customer types and servers induced by the optimal solution of the \gls{LP}~\eqref{eq:LP_eps}.
This property, however, does not hold in general for convex optimization problems \cite{Bertsimas1997}, hence we use \Cref{alg:routing} in \Cref{alg:learning_alg}.

\subsection{Resilience in bursty scenarios}
We consider a day in the data set with relatively many customer arrivals and little available agents, namely December 31, 2001.
We manually mimic three customer surges during office hours by adding artificial customer arrivals: we add 2.000 additional type-2 customer arrivals uniformly distributed between 10:00\AM and 10:10\AM, 2.000 type-3 arrivals between 11:00\AM and 11:10\AM, and 2.000 type-5 arrivals between 12:00\PM and 12:10\PM. The resulting bursty arrival process is shown in \Cref{fig:arrivals_incident_31December2001}. 
Since we found that \gls{UCBQR} can lead to high waiting times, we only consider \gls{UCBQR}--Tree  in this experiment.\\

\paragraph{Delayed recovery of \gls{UCBQR}--Tree after customer surges}
We observe in \Cref{fig:wait_incident_31December2001} that all policies show peaks in customer waiting times after the artificial incidents. 
The benchmark policies \gls{FCFS}---\gls{ALIS}, Greedy, Random, and $\theta\mu$ recover quickly from the customer surges and return to the pre-surge level waiting times at approximately 12:30\PM. 
However, the benchmark policies obtain overall lower payoff than \gls{UCBQR}--Tree, as shown in \Cref{tab:reward_incident_31December2001}. 
The \gls{UCBQR}--Tree policy shows the highest peak in customer waiting time and recovers to pre-surge waiting time levels only at approximately 3\PM. 
This behavior arises because the algorithm makes decisions based only on the arrival and service rate estimators for the current episode, without considering customers who are already waiting in the system. 
Since the surges occur in a small time interval, the arrival rate estimators recover quickly, and the routing scheme is chosen that optimizes the assignment of future customer arrivals, without considering the optimal assignment for the many customers in the system. 
We conclude that \gls{UCBQR}--Tree (and similarly \gls{UCBQR}) are not yet well-suited to handle bursty customer arrival processes, and further research it required. 
To address this, the episode length can be shortened or the values of $\alpha$ and $\beta$ in the arrival rate estimators \eqref{eq:arr_est} can be adjusted to make the routing algorithm more responsive to such customer surges. 
Alternatively, the \gls{LP}~\eqref{eq:LP_eps} can be adapted to account for the number of customers of each type currently in the system. 

\begin{minipage}{.48\textwidth}
\hspace{-1em}
    \centering
    \begin{figure}[H]
        \centering
        \includegraphics[width=\linewidth]{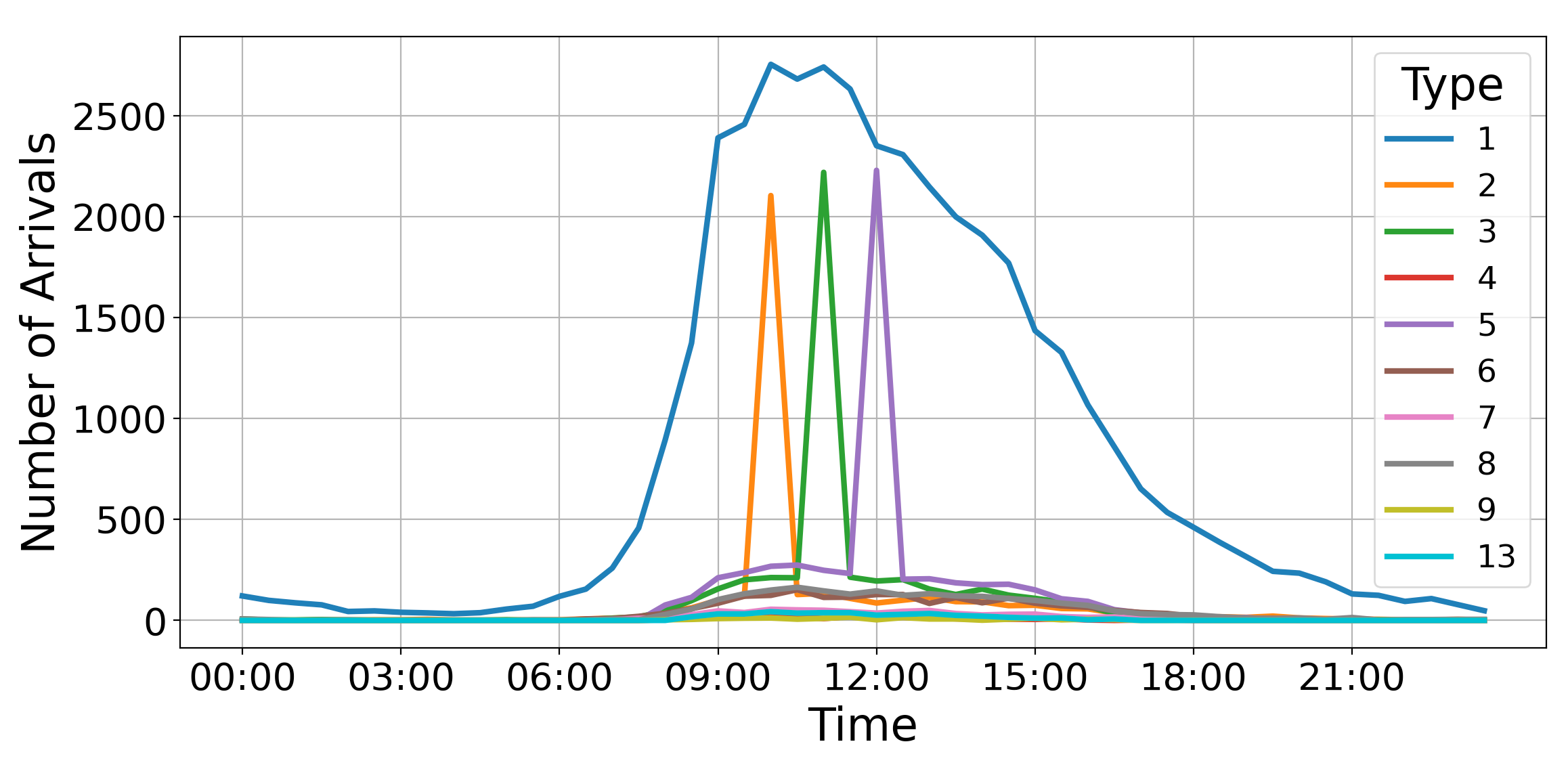}
        \caption{Customer arrivals with addition of the artificial bursty arrival process per 10-minute intervals  for December 31, 2001.}
        \label{fig:arrivals_incident_31December2001}
    \end{figure}
\end{minipage}
\hspace{.1em}
\begin{minipage}{.48\textwidth}
    \centering
    \begin{figure}[H]
        \centering
        \includegraphics[width=\linewidth]{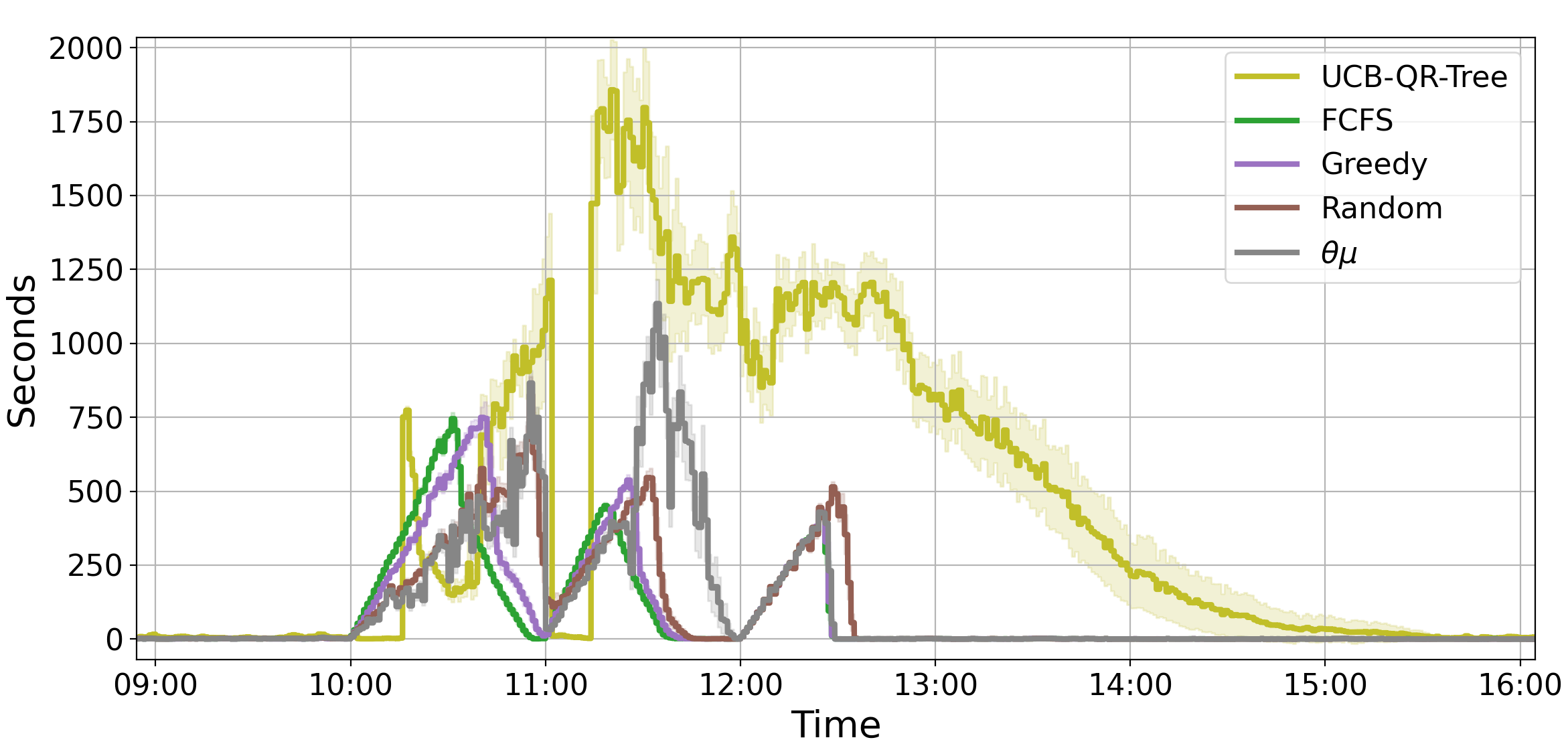}
        \caption{Average customer waiting time for different policies when applied to the modified arrival process on December 31, 2001.}
        \label{fig:wait_incident_31December2001}
    \end{figure}
\end{minipage}
\begin{table}[H]
    \centering
    \footnotesize
    \begin{tabular}{ccccc}
        \toprule
        \gls{UCBQR}--Tree & \gls{FCFS} & Greedy & Random & $\theta\mu$ \\
        \midrule
        0.9910 & 0.9665 & 0.9883 & 0.9662 & 0.9811  \\
        \bottomrule
    \end{tabular}
    \caption{Average cumulative payoff relative to the Oracle policy.}
    \label{tab:reward_incident_31December2001}
\end{table}

\subsection{Robustness against estimation errors}\label{sec:est_error}
We analyze the sensitivity of \gls{UCBQR} to errors in arrival and service rate estimation. 
To this end, we compare four variations of \Cref{alg:learning_alg}. 
The algorithms follow the same steps as described in \Cref{alg:learning_alg}, but differ in the knowledge of system parameters.
\begin{itemize}[nolistsep,noitemsep]
    \item[$(\hat\theta,\hat\lambda,\hat\mu)$] \gls{UCBQR}, which uses \gls{UCB} estimators $\hat\theta$, estimators $\hat\lambda$ in \eqref{eq:arr_est} for the arrival rates, and empirical estimators $\hat\mu$ in~\eqref{eq:serv_est} for the service rates.
    \item[$(\theta,\lambda,\mu)$] Oracle, which uses the true values $\theta,\lambda,\mu$.
    \item[$(\hat\theta,\hat\lambda,\mu)$] \gls{UCBQR}-$\mu$, which uses \gls{UCB} estimators $\hat\theta$ and estimators $\hat\lambda$ in \eqref{eq:arr_est}, but the true service rates $\mu$. 
    \item[$(\hat\theta,\lambda,\hat\mu)$] \gls{UCBQR}-$\lambda$, which uses \gls{UCB} estimators $\hat\theta$, the true arrival rates $\lambda$, and empirical estimators $\hat\mu$ in~\eqref{eq:serv_est} for the service rates. 
\end{itemize}

\paragraph{Estimation error in arrival and service rates has little impact}
We found that both variants \gls{UCBQR}-$\mu$ and \gls{UCBQR}-$\lambda$ perform close to \gls{UCBQR} in terms of payoff accumulation. 
By granting access to the true arrival rate or service rate, the cumulative payoff is only increased in the order of $0.1\%$. 
Next to that, the average waiting times are comparable, as shown in \Cref{fig:wait_est_errors_4June2001}. 
We conclude that the algorithm's performance can only be slightly improved in practice by reducing the estimation errors in the arrival and service rates. 

\begin{minipage}{.4\textwidth}
    \centering
    \begin{figure}[H]
        \centering
        \includegraphics[width=\textwidth]{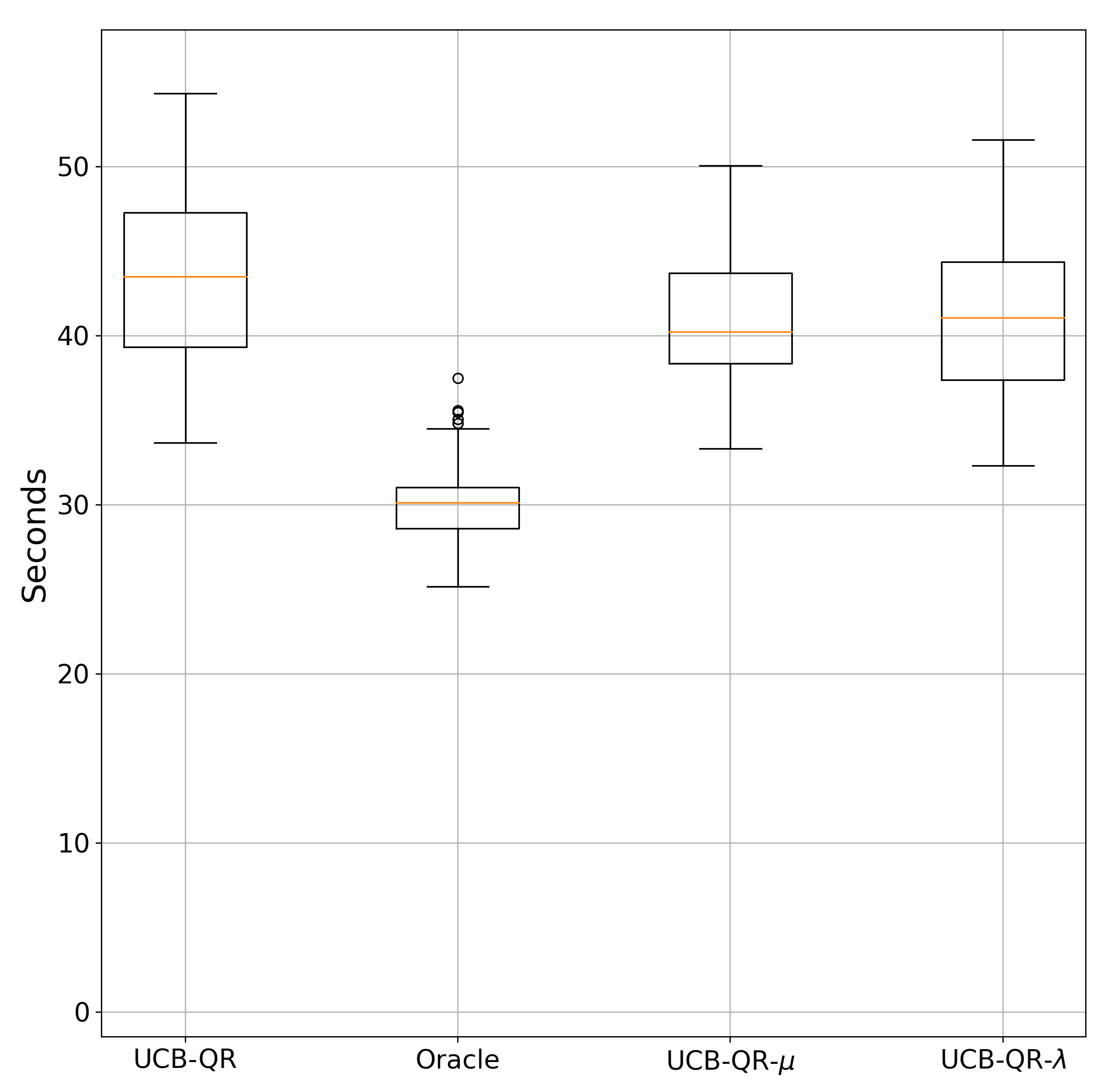}
        \caption{Boxplot of waiting time during office hours for different variants of \gls{UCBQR} relying on different system information when applied to the data of June 4, 2001. }
        \label{fig:wait_est_errors_4June2001}
    \end{figure}
\end{minipage}
\hspace{.1em}
\begin{minipage}{.58\textwidth}
    \centering
    \begin{figure}[H]
        \centering
        \includegraphics[width=\textwidth]{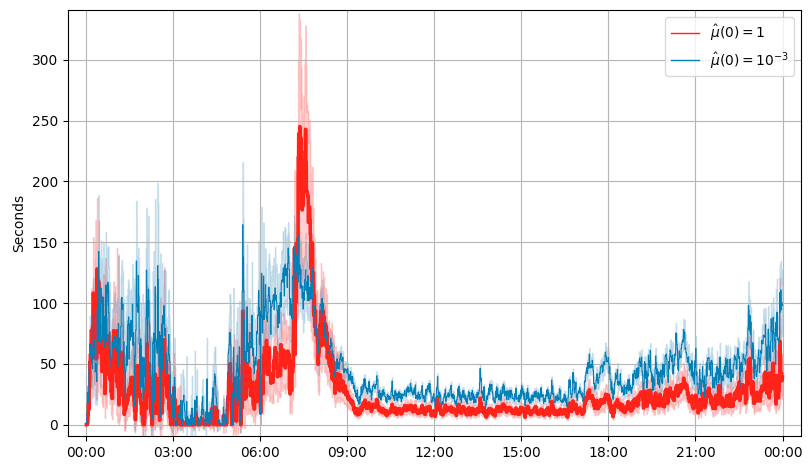}
        \caption{Average waiting time of \gls{UCBQR} for two different initial values of the empirical estimators $\hat\mu$. Overestimation of $\hat\mu$ leads to a significant peak in waiting times around 7\AM. }
        \label{fig:wait_muest_4June2001}
    \end{figure}
\end{minipage}

\paragraph{Additional exploration bonus can lead to system instability}
We note that \gls{UCBQR} uses \gls{UCB} estimators for the average payoff, but empirical estimators for the service rates. 
One might wonder whether adding further exploration bonus to the service rate estimators could improve the algorithm's performance.
In \gls{UCBQR}, the service rate estimators are initialized at a small value ($10^{-3}$), deliberately  underestimating their true mean. 
We find that when $\hat\mu$ overestimates its true mean, the average customer waiting time experiences a significant peak during the start of office hours, as shown in \Cref{fig:wait_muest_4June2001}. 
This can be explained as follows. 
During the low traffic night hours, not all compatible connections have been explored yet, so most estimators $\hat\mu_{ij}$ are still at the initial value at the start of office hours.
The solution of the optimization problem \eqref{eq:LP_eps} is then to route a large amount of customers to one server, much more than its true capacity. 
As a result, the size of one virtual queue is large, while other servers are idling. 
Of course, the empirical estimators are updated after each episode so the algorithm does recover after a few episodes. 
We observe in \Cref{fig:wait_muest_4June2001} that the algorithm with high initial values even performs better in terms of waiting time after the morning peak, since more compatible lines are used that are not explored by the algorithm with small initial values. 
However, since the errors are made during the high traffic morning peak, the overall average waiting time suffers from overestimation of $\hat\mu$: we found that with small initial values, the average waiting time during office hours is approximately 44 seconds, and 58 seconds with large initial values. 
We found that the cumulative payoff is not noticeably affected. \\

To conclude, on this realistic data set, the performance of the \gls{UCBQR} algorithm is not greatly improved by increasing the accuracy of the $\hat\lambda$ and $\hat\mu$ estimators. 
Adding an exploration bonus to the service rate estimators can be beneficial for exploration, but it can lead to an increase in waiting times when the traffic becomes large while not yet all lines have been used.

\subsection{Balancing reactivity and complexity}\label{sec:episode_length}
We analyze the sensitivity of our results with respect to the episode length $h$ used by \gls{UCBQR} in \Cref{line:episode_length}. 
The shorter the episode, the faster the payoff parameters $\theta$ are learned, and the faster the algorithm reacts to changes in arrival rates. 
At the same time, it also increases computational complexity since an optimization problem is solved and the virtual queues are reshuffled each episode (\Cref{line:solve_LP_eps} in \Cref{alg:learning_alg} and \Cref{line:reallocation0}--\ref{line:reallocation2} in \Cref{alg:routing}). \\

\paragraph{Large $h$ leads to significant delays}
We observe in \Cref{tab:runtime_episode_length} that for an episode length exceeding 10 minutes, the episode updates are no longer the dominant factor in computation complexity, since runtime no longer decreases when increasing episode lengths.  
Moreover, the cumulative payoff is not very sensitive to the episode length, where larger episodes give slightly higher payoff since there is less noise in the estimators. 
Customer waiting times, however, are less robust: the average waiting time in \Cref{fig:wait_episode_length_4June2001} increases with the value of $h$. 
Especially the morning and evening hours are chaotic with large peaks in waiting times, since the algorithm is not able to pick up changes in traffic volume adequately fast. 
Hence, $h$ should be chosen small enough to capture changes in time-varying arrival rates.

\begin{minipage}{.77\textwidth}
    \begin{figure}[H]
    \centering
    \includegraphics[width=\linewidth]{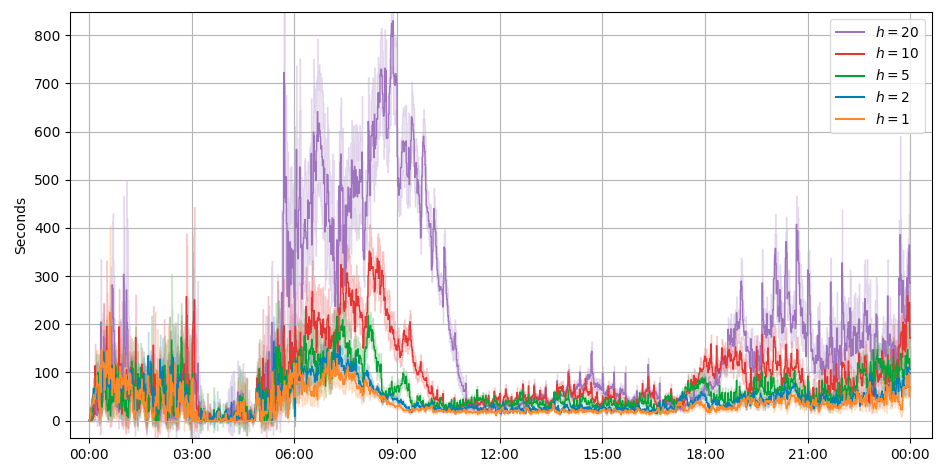}
    \caption{Average customer waiting time of the \gls{UCBQR} algorithm for different values of the episode length $h$ (min) when applied to the data of June 4, 2001. }
    \label{fig:wait_episode_length_4June2001}
\end{figure}    
\end{minipage}
\hspace{.5em}
\begin{minipage}{.2\textwidth}
    \begin{table}[H]
    \centering
    \footnotesize
    \begin{tabular}{cccccc}
    \toprule
    $h$ & Runtime & Payoff \\
    \midrule
    1 & 48 & 0.9914 \\
    2 & 30 & 0.9922\\
    5 & 20 & 0.9931 \\
    10 & 17 & 0.9934 \\
    20 & 17 & 0.9922 \\ 
    \bottomrule
    \end{tabular}
    \caption{Average runtime in seconds and the average cumulative payoff relative to the Oracle policy  of the \gls{UCBQR} algorithm for different values of the episode length $h$ (min) when applied to the data of June 4, 2001. }
    \label{tab:runtime_episode_length}
\end{table}
\end{minipage}

\section{Conclusion}\label{sec:conclusion}
We have demonstrated the effectiveness of the reinforcement learning algorithm \gls{UCBQR} in a practical setting using real-world data, serving as a concrete step towards live implementation of learning-based control policies in real-life skill-based queueing systems. 
We briefly summarize our findings. 
We have shown that \gls{UCBQR} effectively learns the payoff parameters in realistic time-varying environments while its practical computational complexity was comparable with those of benchmark static policies such as \gls{FCFS}---\gls{ALIS} and the $c\mu$ rule. 
Moreover, we have shown that customer waiting times can be drastically reduced with only a minor loss in payoff by using a more efficient heuristic routing rule (\Cref{alg:tree}). 
The algorithm can be easily extended to allow for multiple objectives: we have illustrated by example the joint optimization of payoff accumulation and server fairness, where a tuning parameter is used to balance the trade--off. 
In case of sudden customer surges, we observed that \gls{UCBQR} does not recover to the pre-surge performance as quickly as benchmark policies. 
To improve robustness against such bursty arrival processes however, better finetuning of the algorithm parameters is required.
We found that for this data set, not much can be gained in terms of payoff or customer delay by reducing the noise in arrival and service rate estimators.
Lastly, we have shown that the episode length as parameter of the \gls{UCBQR} algorithm requires careful tuning depending on the application to maintain short customer delays. \\

For this study we used a public data set of call center logs from 2001 to 2003, although a case study for different applications, such as data centers or cloud computing networks, with more recent data would be useful follow-up research for practice.
Other possible directions for future research include considering applications with customer abandonments, increasing adaptability of the algorithm by using event-based updates rather than episodes, and generalizing for multi-pool systems.

\subsection*{Acknowledgments}
This work is part of \emph{Valuable AI}, a research collaboration between the Eindhoven University of Technology and the Koninklijke KPN N.V.
Parts of this research have been funded by the EAISI's IMPULS program, and by Holland High Tech | TKI HSTM via the PPS allowance scheme for public-private partnerships.

\newpage
\bibliographystyle{unsrt}
\bibliography{Literature}

\newpage
\appendix
\section{Data and simulation description}\label{app:implementation}

\begin{figure}[H]
    \centering
    \includegraphics[width=\linewidth]{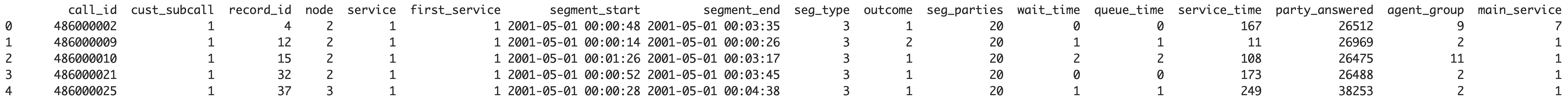}
    \caption{
    Excerpt of the call log table of the data set. 
    Here, ``outcome'' encodes the reason why the call ended, such as abandonment or transfer, and ``party\_answered'' is the id of the agent that answered the call. }
    \label{fig:call_log_1May2001}
\end{figure}
The data set contains 130 gigabytes of tables with detailed information about call logs and agent shifts from a U.S.\ bank call center over the period from April 2001 to October 2003. 
An example of such call logs is shown in \Cref{fig:call_log_1May2001}.
The call center operates 24 hours a day, 7 days a week. 
On an average weekday, approximately 1.200 agents handle around 50.000 calls. 
Calling customers are processed by a \gls{VRU} which determines the type of service requested by the customer. 
We label the different service requests as customer types.
Not all customer types appear in each month of the data. 
In May, June, and November 2002, new services were introduced by the call center's management, bringing the total to 17 different customer types. \\

Service in the call center of the data set is skill-based.
The agents are partitioned into 47 different groups based on their service capabilities. 
The data set contains detailed information about the agents' work hours, calling times, and idle times.

\subsection{Data preparation}
Since the service categories and the agent populations change over time, we consider the compatibility between customer types and agent groups per month. 
For a given month, we define a customer type and agent group to be \emph{compatible} if there are at least 100 calls of that combination within the month. 
An example of the customer types, agent groups and their compatibility is given in \ref{app:input_data_June2001}.
In the data set, at most 1\% of the customers abandon the system before speaking to an agent. 
We neglect these relatively rare abandonments and only consider calls where the customer interacts with an agent. 
Within one month, about 3\% of the customers interact with more than one agent, either due to call transfers or multiple calls initiated by the customer.
For simplicity, we neglect this aspect and treat each call independently as a unique customer. 

\subsection{Simulation description}
For this study we created a discrete-event simulation to model a general multi-class, multi-server queueing system with customer--server compatibility constraints.
In short, the simulation framework is described as follows. \\

{\bf Input.}
To run the simulation, one has to specify the queueing system's configuration including the number of customer types, the number of servers, and the compatibility relations between customers and servers.
For each customer type an efficient list structure (deque) is created. 
Next to that, we create a Server object for each server that stores its id number and status (idle, busy, or inactive). 
For each compatible customer--server pair,  a payoff distribution has to be specified.
For the arrival process, either a list of customer arrival times or an interarrival distribution is required. 
For the service distributions, the simulation supports both server dependent and customer--server dependent distributions. 
Optionally, an agent schedule can be provided that specifies the number of active agents at certain timestamps. \\

{\bf Simulation dynamics.}
The simulation follows a discrete-event approach~\cite{Cassandras2008}: future events are stored in an event queue and the simulation progresses by repeatedly processing the next scheduled event.
Events can be of the type arrival, departure, agent schedule update, or episode end. 
Upon an arrival, the customer is allocated to either an idle server, or a (virtual) queue according to the routing policy. 
Upon a service completion, a payoff is sampled according to the payoff distribution of the customer--server pair, empirical estimators for the service duration and average payoff are updated, and the next service completion is scheduled if there are customers in the queues allocated to this server. Otherwise, the server status is set to idle. \\

{\bf Simulation output.}
The simulation output is an event log containing details (time stamp, customer type and id, server id) of all events including arrivals, the start of services, and service completions. 
From the event log, key performance indicators such as average payoff, customer waiting times, server utilization, and queue lengths are obtained. 
Each simulation is ran multiple times to reduce the fluctuations in the randomly sampled arrival times (unless the arrivals are specified according to a deterministic list), service times, and payoffs.

\section{Input data of the first week of June 2001}\label{app:input_data_June2001}
\begin{figure}[H]
    \centering
    \includegraphics[width=\textwidth]{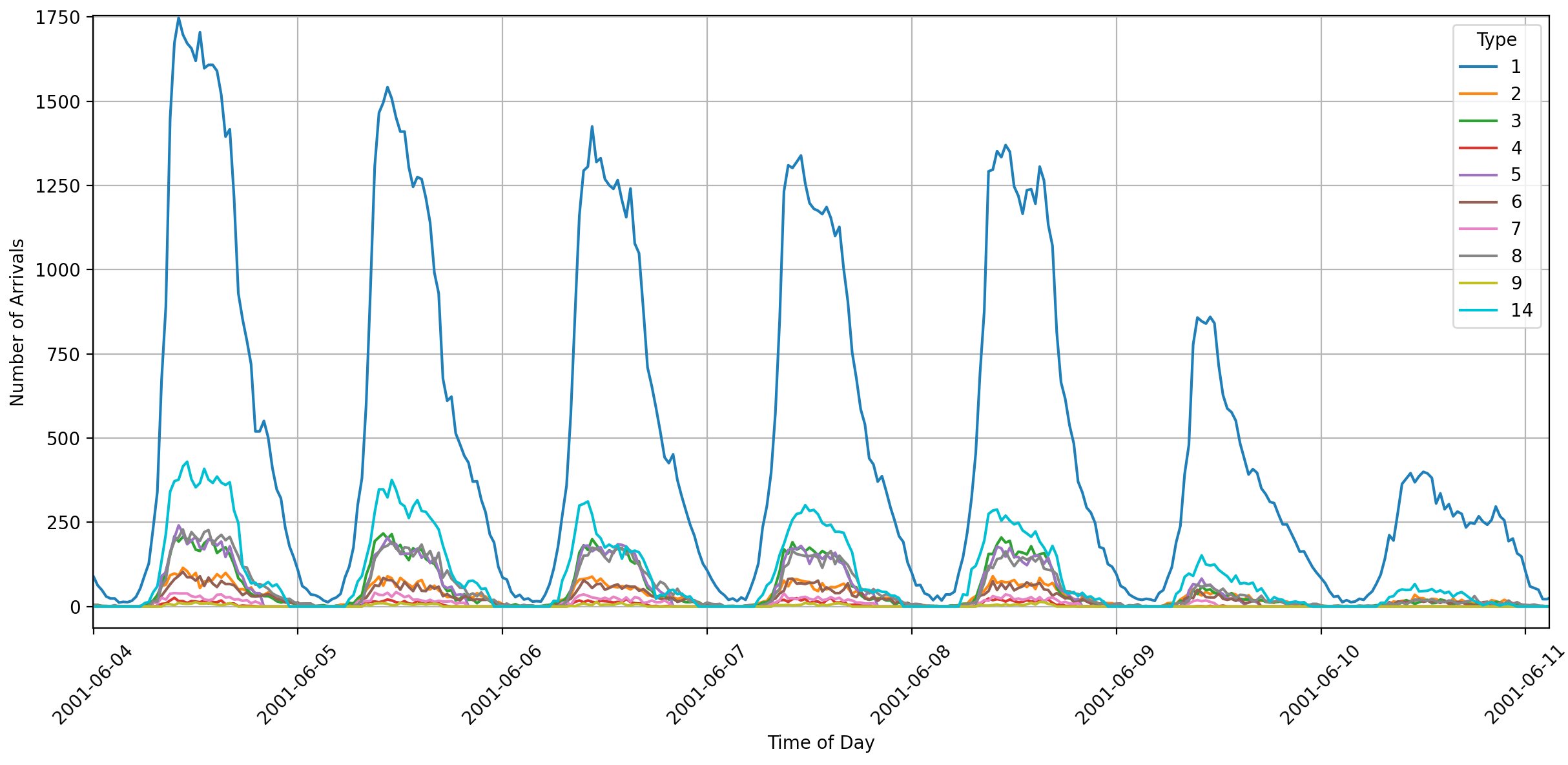}
    \caption{Number of customer arrivals per type accumulated per 30-minute intervals for the first week of June 2001. }
    \label{fig:arrivals_4June2001}
\end{figure}
\begin{figure}[H]
    \centering
    \includegraphics[width=\textwidth]{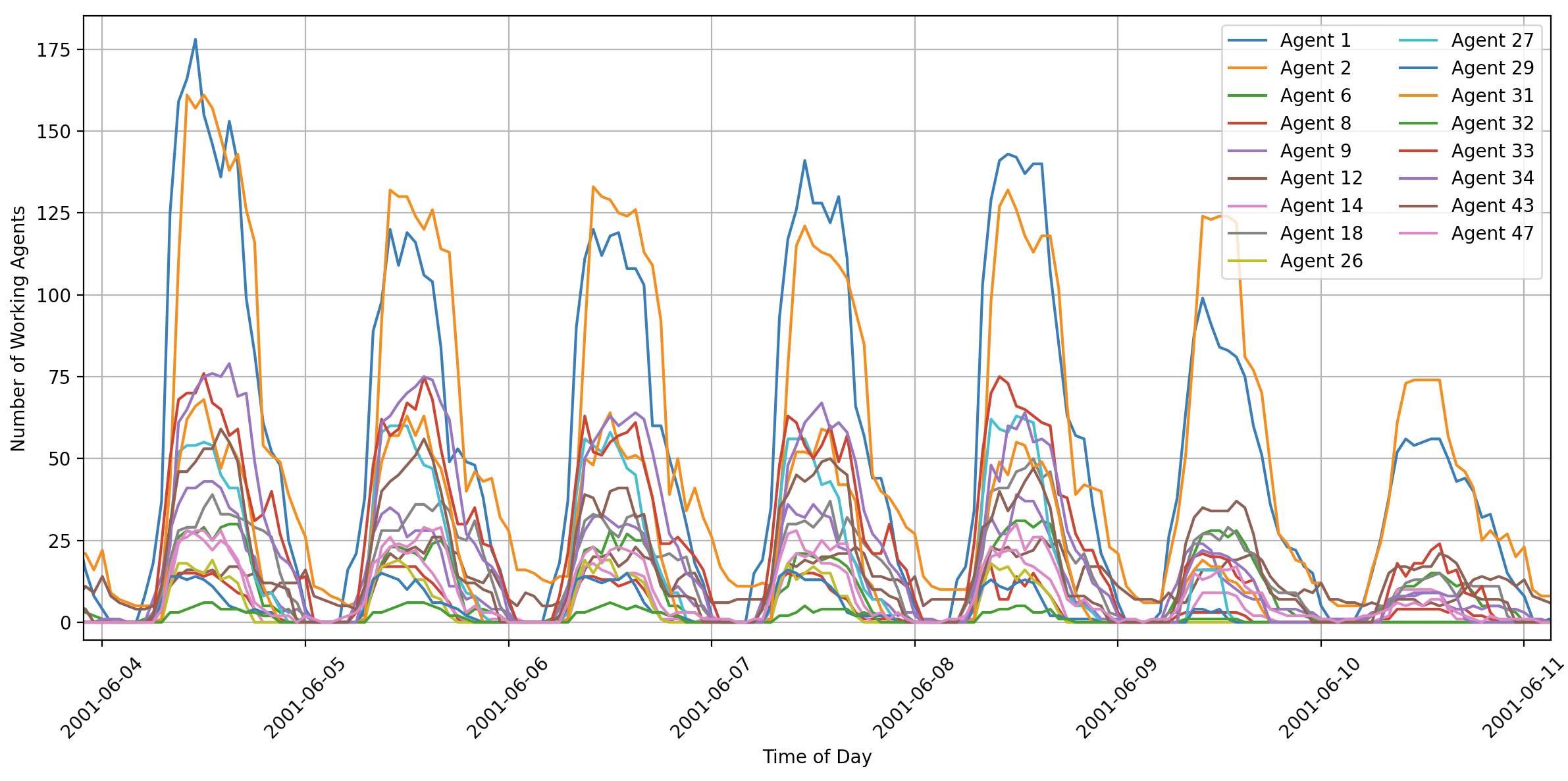}
    \caption{Number of working agents per group accumulated per hour for the first week of June 2001.
    Here, working means being available to answer calls. Most agents are active between 6\AM and 9\PM.}
    \label{fig:agent_schedule_4June2001}
\end{figure}

\begin{table}[H]
    \centering
    \begin{tabular}{c|cccccccccc} 
    \toprule
    \diagbox{\scriptsize{Agent}}{\scriptsize{Customer}} &  1 & 2 & 3 & 4 & 5 & 6 & 7 & 8 & 9 & 14 \\ 
    \midrule
    1 & 0.84 &  &  &  &  &  &  &  &  &  \\ 
    2 & 0.84 & 0.82 & 0.93 &  &  &  &  & 0.67 &  &  \\ 
    6 & 0.81 &  & 1.00 &  &  &  &  &  &  &  \\ 
    8 & 0.84 &  &  &  &  & 0.71 &  &  &  &  \\ 
    9 & 0.87 &  & 0.89 &  &  &  & 0.97 &  &  &  \\ 
    12 & 0.85 & 0.89 & 0.84 &  &  &  &  & 0.61 &  &  \\ 
    14 & 0.80 &  &  &  &  &  &  &  & 0.69 &  \\ 
    18 &  & 0.86 &  &  &  &  &  &  &  &  \\ 
    26 &  &  & 0.90 &  &  &  &  &  &  &  \\ 
    27 &  & 1.00 & 0.90 & 1.00 &  &  &  &  &  &  \\ 
    29 &  &  & 0.91 & 0.91 &  &  &  &  &  &  \\ 
    31 & 0.88 &  &  &  & 0.84 &  &  &  &  &  \\ 
    32 &  &  &  &  & 0.84 & 0.69 &  &  & 0.90 &  \\ 
    33 &  &  &  &  &  & 0.86 &  &  &  &  \\ 
    34 &  &  &  &  &  &  &  & 0.78 &  &  \\ 
    43 &  &  &  &  &  &  &  &  &  & 0.95 \\ 
    47 & 0.92 &  &  &  &  &  &  &  &  & 0.96 \\ 
    \bottomrule
    \end{tabular}
    \caption{True value of the payoff parameter $\theta_{ij}$ for all lines $(ij)\in\calL$ for 4 June 2001.
    $\theta_{ij}$ is the fraction of successful $(ij)$ calls in the data set on this day, i.e., calls that are not transferred nor end in a conference. 
    An empty cell means that the customer--agent pair is incompatible.  }
    \label{tab:true_theta_4June2001}
\end{table}

\section{Adapted payoff value}\label{app:adapted_theta}
We consider the same setting as \Cref{sec:experiment_initial}, but use an adapted payoff parameter $\theta'$.
We normalize the values $\theta_{ij}$ to the range $[0,1]$ and apply a power transformation to increase the relative differences. 
Concretely, we define
\begin{align}\label{eq:adapted_theta}
    \theta_{ij}' &= 
        \sqrt{\frac{\theta_{ij}}{\max_k \theta_{ik} - \min_\ell \theta_{i\ell}}},
\end{align}
for customer types $i\in\calI$ with more than one compatible line, and $\theta_{ij}' = \sqrt{\theta_{ij}}$ for customer types with only one connection. 
For example, the adapted value $\theta'$ for June 4 2001 is shown in \Cref{tab:adapted_theta_4June2001}, which shows much more variation than the original value of $\theta$ (\Cref{tab:true_theta_4June2001} in \ref{app:input_data_June2001}).
\begin{table}[H]
\centering
\begin{tabular}{c|cccccccccc}
\toprule
\diagbox{\scriptsize{Agent}}{\scriptsize{Customer}} & 1 & 2 & 3 & 4 & 5 & 6 & 7 & 8 & 9 & 14 \\ 
\midrule
1 & 0.60 &  &  &  &  &  &  &  &  &  \\ 
2 & 0.59 & 0.00 & 0.76 &  &  &  &  & 0.59 &  &  \\ 
6 & 0.37 &  & 1.00 &  &  &  &  &  &  &  \\ 
8 & 0.58 &  &  &  &  & 0.40 &  &  &  &  \\ 
9 & 0.76 &  & 0.54 &  &  &  & 0.98 &  &  &  \\ 
12 & 0.69 & 0.61 & 0.00 &  &  &  &  & 0.00 &  &  \\ 
14 & 0.00 &  &  &  &  &  &  &  & 0.00 &  \\ 
18 &  & 0.45 &  &  &  &  &  &  &  &  \\ 
26 &  &  & 0.61 &  &  &  &  &  &  &  \\ 
27 &  & 1.00 & 0.58 & 1.00 &  &  &  &  &  &  \\ 
29 &  &  & 0.65 & 0.00 &  &  &  &  &  &  \\ 
31 & 0.84 &  &  &  & 1.00 &  &  &  &  &  \\ 
32 &  &  &  &  & 0.00 & 0.00 &  &  & 1.00 &  \\ 
33 &  &  &  &  &  & 1.00 &  &  &  &  \\ 
34 &  &  &  &  &  &  &  & 1.00 &  &  \\ 
43 &  &  &  &  &  &  &  &  &  & 0.00 \\ 
47 & 1.00 &  &  &  &  &  &  &  &  & 1.00 \\ 
\bottomrule
\end{tabular}
\label{tab:adapted_theta_4June2001}
\caption{Adapted value of $\theta'$ according to~\eqref{eq:adapted_theta} for 4 June 2001.}
\end{table}

\begin{figure}[H]
    \centering
    \includegraphics[width=1\linewidth]{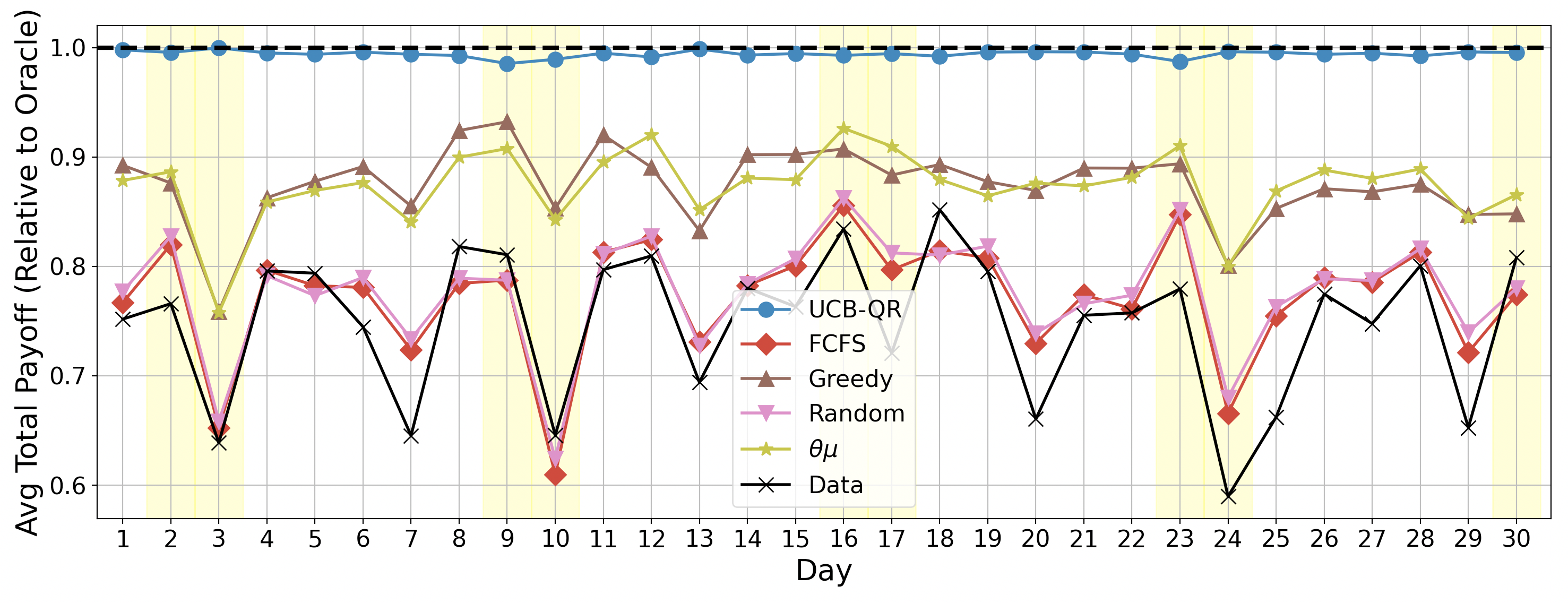}
    \caption{Average total payoff  per day in June 2001, expressed relative to the Oracle policy, based on the adapted payoff parameter $\theta'$ in \eqref{eq:adapted_theta}.}
    \label{fig:reward_adapted_theta_June2001}
\end{figure}

Consequently, we run the same simulation as described in \Cref{sec:experiment_initial}, but now using $\theta'$.
The average payoff is shown in \Cref{fig:reward_adapted_theta_June2001}.

\newpage
\section{Performance of \Cref{alg:tree} on a small example}\label{app:routing_q_small}
Consider a queueing system with $I=3$ customer types and $J=3$ servers, and full compatibility $\calL= \{(i,j) \ \forall i\in\calI,j\in\calJ\}$ as shown in \Cref{fig:graph_q_small}.
Arrivals of type-$i$ occur according to independent Poisson processes with rate $\lambda_i$ and type-$(ij)$ service times are independent and exponentially distributed with rate $\mu_{ij} = \mu_j$ for all $i\in\calI$ and $j\in\calJ$. \\

We consider the following non-degenerate \gls{BFS} of \eqref{eq:LP_eps},
\begin{subequations}\label{eq:sol_q_small}
\begin{align}
    x_{11} &= \mu_1-\eps = 1-\eps, \\
    x_{12} &= \lambda_1-x_{11} = 2+\eps, \\
    x_{22} &= \mu_2(1-\eps) - x_{12} = 3-6\eps, \\
    x_{23} &= \lambda_2 - x_{22} = 4+6\eps, \\
    x_{32} &= \lambda_3 = 5. 
\end{align}
\end{subequations}

In this case, server 3 has slack, since \eqref{eq:LP_eps_constr_mu} is not binding. 
The spanning forest induced by $x$ is shown in \Cref{fig:sf_q_small}. 

\begin{minipage}{.45\textwidth}
    \centering
    \begin{figure}[H]
        \centering
            \begin{tikzpicture}[
                server/.style={circle, minimum size = .9cm, thick,draw},
                scale=.7
                ]
                \foreach \x/\y in {1/3,2/7,3/5}{ 
                    \node[draw = none] at (2*\x,0) (queue\x) {};
                    \node[above of = queue\x, yshift = -.5cm] (labelqueue\x) {};
                    \draw[thick] (queue\x.center) --++(-.5,0) --++(0,1.2); 
                    \draw[thick] (queue\x.center) --++(.5,0) --++(0,1.2); 
                    \node[draw = none, above of = queue\x, yshift = .2cm] {$\lambda_\x = \y$};
                }
                \foreach \x/\y in {1/1, 2/5, 3/10}{
                    \node[server] at (2*\x,-2.5) (server\x) {};
                    \node[below of = server\x, yshift=.1cm] {$\mu_\x = \y$};
                }
                \draw[ultra thick, black] (queue1.center) -- (server1.north);
                \draw[ultra thick, black] (queue1.center) -- (server2.north);
                \draw[ black] (queue1.center) -- (server3.north);
                \draw[black] (queue2.center) -- (server1.north);
                \draw[ultra thick, black] (queue2.center) -- (server2.north);
                \draw[ultra thick, black] (queue2.center) -- (server3.north);
                \draw[black] (queue3.center) -- (server1.north);
                \draw[black] (queue3.center) -- (server2.north);
                \draw[ultra thick, black] (queue3.center) -- (server3.north);
            \end{tikzpicture}
            \caption{Queueing system with full compatibility. The lines with non-zero mass in $x$ in \eqref{eq:sol_q_small} are shown in bold. }
            \label{fig:graph_q_small}
        \end{figure}
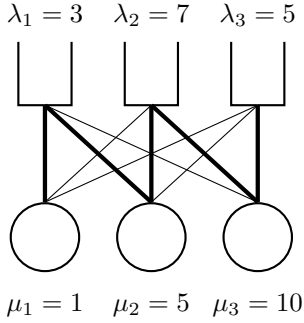
\end{minipage}
\hspace{3em}
\begin{minipage}{.45\textwidth}
    \centering
    \begin{figure}[H]
        \centering
        \begin{tikzpicture}[
            scale = .7,
                server/.style={circle, minimum size = .6cm, thick,draw},
                q/.style={rectangle, minimum size = .6cm, thick,draw}
                ]
                \node[server] at (0,0) (server3) {};
                \node at (server3) {$3$};
                \node[q, below of = server3, xshift = -.7cm, yshift=0cm] (queue2) {2};
                \node[q, below of = server3, xshift = .7cm, yshift=0cm] (queue3) {3};
                \node[server, below of = queue2, yshift=0cm] (server2) {2};
                \node[q, below of = server2, yshift=0cm] (queue1) {1};
                \node[server, below of = queue1, yshift=0cm] (server1) {1};
                \draw[thick] (server3.south) -- (queue2.north);
                \draw[thick] (server3.south) -- (queue3.north);
                \draw[thick] (queue2.south) -- (server2.north);
                \draw[thick] (server2.south) -- (queue1.north);
                \draw[thick] (queue1.south) -- (server1.north);
        \end{tikzpicture}
        \caption{Spanning forest induced by $x$ in \eqref{eq:sol_q_small}. Here, customer types are represented by squares and servers by circles. Server 3 is the root since it has slack. }
        \label{fig:sf_q_small}
    \end{figure}
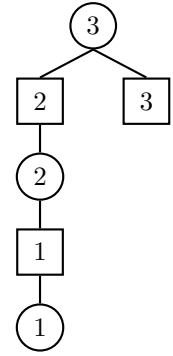
\end{minipage}

We apply the routing rule \Cref{alg:tree} to route customers according to the rates \eqref{eq:sol_q_small}. 
The results are based on 100 independent simulations with a runtime of 500 time units, and episode lengths $h=10$.
\Cref{fig:rates_q_small} shows the running average empirical routing rates.
After 500 time units, the routing rates have converged to approximately
\begin{align}
    \frac{D_{11}(t)}{t} &\approx 0.863, \ 
    \frac{D_{12}(t)}{t} \approx 2.153, \ 
    \frac{D_{22}(t)}{t} \approx 2.649, \ 
    \frac{D_{23}(t)}{t} \approx 4.338, \ 
    \frac{D_{33}(t)}{t} \approx 5.  
\end{align}
There does not exist a value of $\eps > 0$ such that the empirical routing rates agree with the analytical solution \eqref{eq:sol_q_small}. Hence, the routing rates converge approximately, but not exactly to the target $x$. \\

Next to that, the server loads according to the analytical solution \eqref{eq:sol_q_small} are given by $\rho_1 = x_{11}/\mu_1 = 1-\eps$, $\rho_2 = (x_{12}+x_{22})/\mu_2 = 1-\eps$, and $\rho_3 = (x_{23}+x_{33})/\mu_3 = (9+6\eps)/10$. 
However, we find that the empirical server loads converge to approximately (see \Cref{fig:working_q_small})
\begin{align}
    \frac{W_1(t)}{t} &\approx 0.852, \ 
    \frac{W_2(t)}{t} \approx 0.955, \ 
    \frac{W_3(t)}{t} \approx 0.936, 
\end{align}
where $W_j(t)$ is the total amount of time server $j$ has been working (not idling) up to time $t$. 
This does not match the analytical solution $\rho$, since e.g.\ $\rho_1=\rho_2$ but $W_1(t)/t \neq W_2(t)/t$.

\begin{minipage}{.47\textwidth}
    \begin{figure}[H]
        \includegraphics[width=1\textwidth]{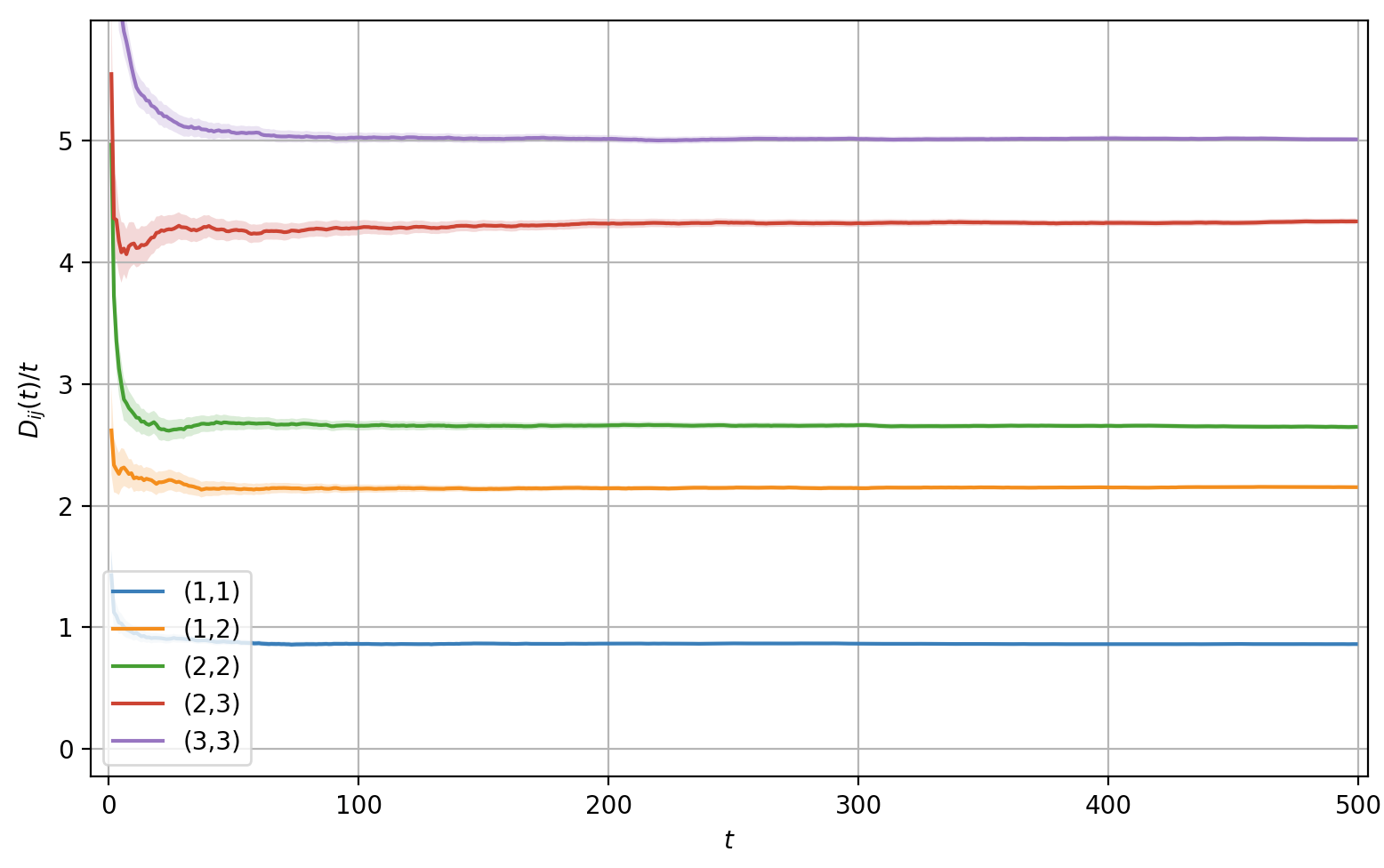}
        \caption{Empirical routing rates $D_{ij}(t)/t$ over time $t$ for the different lines $(ij)$.}
        \label{fig:rates_q_small}
    \end{figure}
\end{minipage}
\hspace{.2em}
\begin{minipage}{.47\textwidth}
    \centering
    \begin{figure}[H]
        \centering
        \includegraphics[width=1\textwidth]{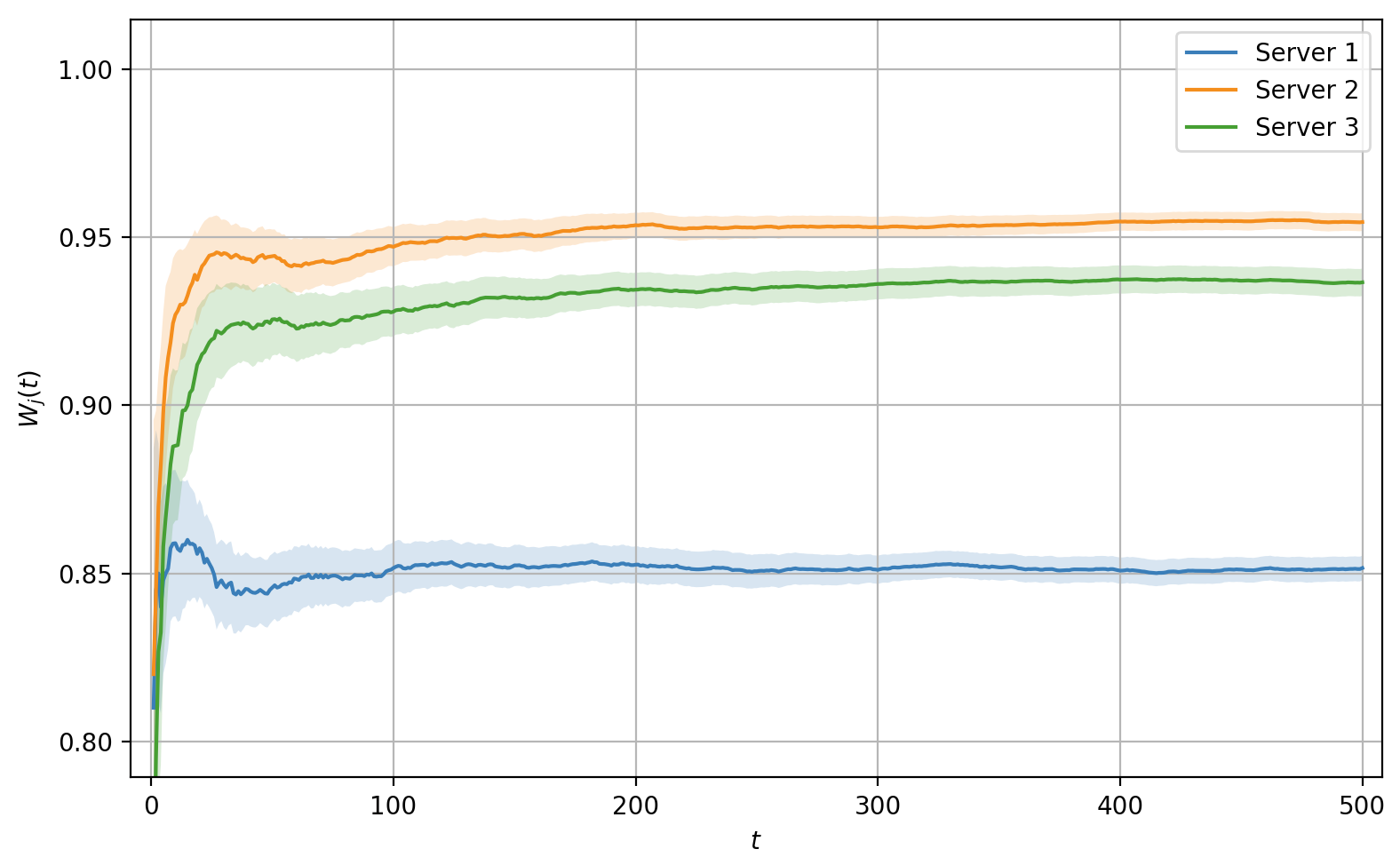}
        \caption{Empirical server load $W_j(t)/t$. }
        \label{fig:working_q_small}
    \end{figure}
\end{minipage}

To analyze the difference in performance between the routing rules \Cref{alg:routing} and \Cref{alg:tree}, we run the \gls{UCBQR}, \gls{UCBQR}-Tree, and the Oracle policy for the queueing system as shown in \Cref{fig:graph_q_small}. 
The cumulative payoff is shown in \Cref{fig:reward_q_small} and  the average customer waiting time in \Cref{fig:wait_q_small}.
The \gls{UCBQR} and \gls{UCBQR}-Tree algorithms obtain a very similar payoff rate.
For both \gls{UCBQR} and the Oracle policy, customers waiting times exhibit sharp peaks at new episodes, i.e., every $h=10$ time units. 
This is an effect of \Cref{alg:routing}: customers are redistributed amongst the virtual queues at the start of each new episode, and therefore customers that have been waiting for some time are suddenly at the head of the virtual queue and start service. 
The waiting times under \gls{UCBQR}-Tree are significantly lower than those of \gls{UCBQR} and the Oracle policy, showing that \Cref{alg:tree} is more efficient in customer routing than \Cref{alg:routing}, with only a slight loss in payoff. 

\begin{minipage}{.35\textwidth}
    \centering
    \begin{figure}[H]
        \centering
        \includegraphics[width=1\textwidth]{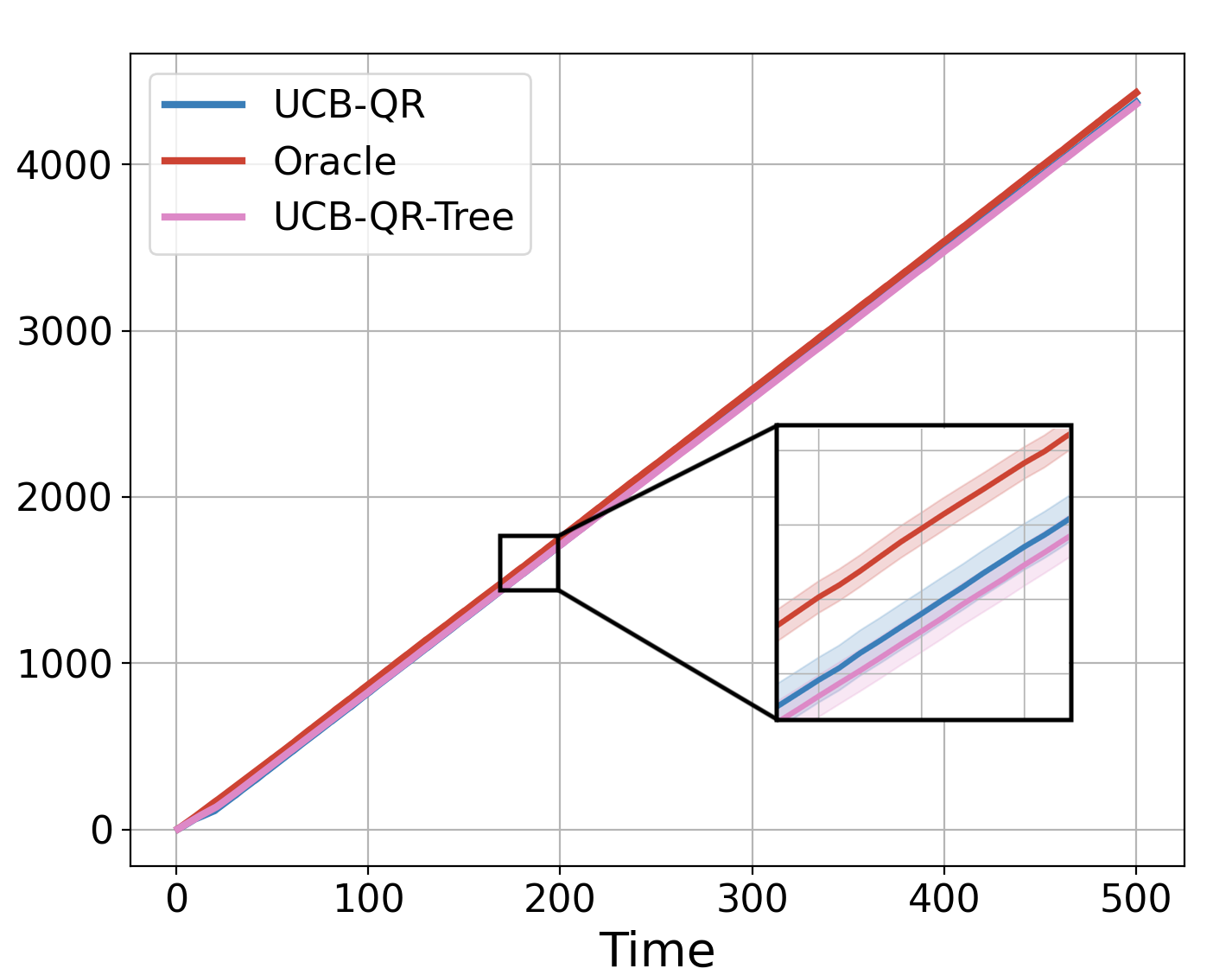}
        \caption{Cumulative payoff.}
        \label{fig:reward_q_small}
    \end{figure}
\end{minipage}
\begin{minipage}{.64\textwidth}
    \centering
    \begin{figure}[H]
        \centering
        \includegraphics[width=1\textwidth]{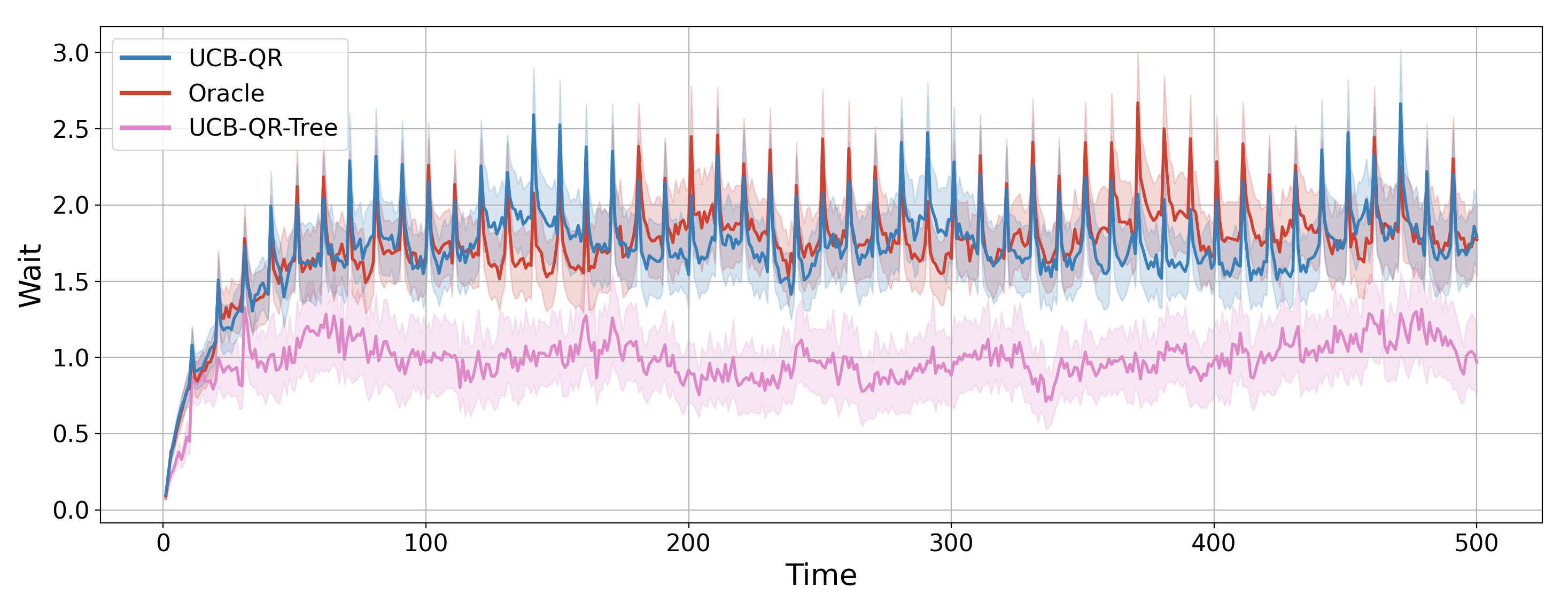}
        \caption{Average customer waiting time.}
        \label{fig:wait_q_small}
    \end{figure}
\end{minipage}

\end{document}